\documentclass[sigconf,screen,nonacm]{acmart}

\usepackage{algorithm}
\usepackage{algorithmicx}
\usepackage[commentColor=gray!90!black,beginComment=$\triangleright$~,beginLComment=$\triangleright$~,endLComment=]{algpseudocodex}

\usepackage{stfloats}
\usepackage{url}
\usepackage{enumitem}
\usepackage{multirow}
\usepackage{booktabs}
\usepackage{threeparttable}
\usepackage{bbding}
\usepackage{arydshln}
\usepackage{makecell}
\usepackage{xspace}
\usepackage{subcaption}

\newcommand{\todo}[1]{\textcolor{black}{#1}}
\newcommand \tool {\textsc{Venom}\xspace}

\usepackage{microtype}
\begin{document}

\title{A general approach to enhance the survivability of backdoor attacks by decision path coupling}

\author{Yufei Zhao}
\affiliation{
  \institution{Fudan University}
  \city{Shanghai}
  \country{China}}

\author{Dingji Wang}
\affiliation{
    \institution{Fudan University}
    \city{Shanghai}
    \country{China}}

\author{Bihuan Chen}
\authornote{Corresponding Author.}
\affiliation{
    \institution{Fudan University}
    \city{Shanghai}
    \country{China}}

\author{Ziqian Chen}
\affiliation{
    \institution{Fudan University}
    \city{Shanghai}
    \country{China}}

\author{Xin Peng}
\affiliation{
    \institution{Fudan University}
    \city{Shanghai}
    \country{China}}

\begin{abstract}
Backdoor attacks have been one~of the emerging security threats~to deep neural networks (DNNs), leading to serious consequences.~One of the mainstream backdoor defenses is model reconstruction-based. Such defenses adopt model unlearning or pruning~to~eliminate~backdoors. However, little attention has been~paid~to survive from such defenses. To bridge the gap, we propose \tool,~the~first~generic~backdoor attack enhancer to improve the survivability~of~existing backdoor attacks against model reconstruction-based defenses. We formalize \tool as a binary-task optimization problem. The first~is~the original backdoor attack task to preserve the original attack capability, while the second is the attack enhancement task to improve the attack survivability. To realize the second task, we propose attention imitation loss to force the decision path of poisoned samples in backdoored models~to~couple~with the crucial decision path of benign samples, which makes backdoors difficult to eliminate.~Our extensive evaluation on two DNNs and three datasets has demonstrated that \tool~significantly improves the survivability of eight state-of-the-art attacks against~eight state-of-the-art defenses without impacting the capability of the original attacks.
\end{abstract}

\maketitle

\section{Introduction}\label{sec:intro}

Deep neural networks (DNNs) have been successfully adopted~in~a variety of domains, including facial recognition~\cite{tang2004frame,parkhi2015deep}, natural~language processing~\cite{devlin2018bert}, 
and autonomous driving~\cite{grigorescu2020survey}. However, it often requires a large amount of computing resources to train DNNs. To reduce the training cost, it is a common practice for users to directly use models published to model repositories 
(e.g., Hugging Face), or outsource the model training process to cloud computing platforms (e.g., Google Cloud). While it is cost-efficient to adopt third-party models or third-party platforms, it does introduce security threats.

Backdoor attack has become one of the emerging security threats. Malicious model publishers and untrusted platforms may inject~backdoors into DNNs during the training process by poisoning a portion of training samples~\cite{li2022backdoor}. The backdoored DNNs misclassify samples with attacker-specified trigger patterns into target labels, while~exhibiting normal behavior on benign samples. Many attacks also try to improve the stealthiness of triggers (e.g.,~\cite{turner2019label,nguyen2020input,li2021invisible,nguyen2021wanet}).~Such~attacks can lead to serious consequences. For example, a backdoored autonomous driving system could classify a stop sign with a particular sticker posted on it as a speed limit sign~\cite{gu2019badnets},~leading~to~erroneous decision-making and threatening driver~safety. In this work, we focus on image classification, the most popular attacking target.

To mitigate backdoor threats in DNNs, there are two categories~of mainstream defenses. \textit{Data distribution-based defenses}~(e.g., \cite{tran2018spectral, chen2018detecting})~assume that models trained~on~poisoned datasets tend to learn separable latent representations~for~poisoned and benign samples.~Therefore, they identify and eliminate~poisoned samples via cluster analysis in the latent space. Instead of focusing on data distribution,~\textit{model reconstruction-based defenses} (e.g., \cite{liu2018fine, zheng2022pre}) concentrate on model~behavior. They are built upon two assumptions, (i) poisoned samples exhibit different activation values in backdoored models~from~benign samples, and (ii) poisoned samples occupy extra decision paths in backdoored models than benign samples. Therefore, they leverage model unlearning or model pruning to remove backdoors. 

In response to advances in defense techniques, previous attacks have achieved progress in improving latent inseparability~\cite{shokri2020bypassing,ijcai2022p0242,xia2022enhancing,qi2022revisiting} to bypass data distribution-based defenses. However,~little~attention has been given to survive from model reconstruction-based defenses. To fill this gap, we propose~a \textit{generic} backdoor~attack~enhancer, \tool, which enhances~existing attacks~by~improving their \textit{survivability} against existing model reconstruction-based~defenses. In general, the key idea of \tool is to break the~two assumptions that those defenses are built upon, making those defenses~not~applicable. When equipped with \tool, existing attacks not only~preserve their original capabilities (e.g., trigger stealthiness and latent inseparability), but also maintain high attack success rate after~those defenses are applied, or in the worst case at least cause a significant decrease in benign accuracy, making the model unusable.

We formulate \tool as a binary-task optimization problem.~The first task is the original backdoor attack task, which aims to preserve the original attack capability, while the second task~is~the~attack~enhancement task, which aims to enhance the attack survivability.~To realize the second task, we first generate the decision path (a set~of neurons) that is crucial for classifying benign samples from a clean model. Then, we propose attention imitation loss to force the decision path of poisoned samples in backdoored models~to~couple~with the generated crucial decision path of benign samples. The coupling is reflected in two dimensions. First, the decision path of poisoned samples overlap with that of benign samples, which breaks~the~assumption (ii). Second, the activation behavior of poisoned samples~is similar to that of benign samples, which breaks the assumption~(i). As a result, the backdoor becomes~difficult~for~defense techniques to eliminate. Even if certain defense technique successfully eliminates the backdoor, it will inevitably lead to a significant decrease in the classification accuracy of benign samples.

We conduct extensive experiments with two widely used DNN structures (i.e., VGG19-BN~\cite{simonyan2015very} and PreActResNet18~\cite{he2016identity}) and three popular datasets (i.e., CIFAR-10~\cite{krizhevsky2009learning},  CIFAR-100~\cite{krizhevsky2009learning} and GTSRB~\cite{6706807}). We use VGG19-BN for all the three datasets, and PreActResNet18 for the CIFAR-10 dataset. We evaluate eight state-of-the-art backdoor attacks, i.e., BadNets~\cite{gu2019badnets}, Blend~\cite{chen2017targeted}, TrojanNN~\cite{liu2018trojaning}, LC~\cite{turner2019label},~SSBA \cite{li2021invisible}, Inputaware~\cite{nguyen2020input}, WaNet~\cite{nguyen2021wanet} and Adap-Blend~\cite{qi2022revisiting},~and~eight state-of-the-art backdoor defenses, i.e., FT~\cite{liu2018fine}, NAD~\cite{li2021neural}, I-BAU~\cite{zeng2022adversarial},  NC~\cite{wang2019neural}, BNP~\cite{zheng2022pre}, 
FP~\cite{liu2018fine}, CLP~\cite{zheng2022data}, NPD ~\cite{zhu2023neural}.

We first measure how \tool affects the capability of the original attacks when no defense is applied. On average, \tool slightly~improves the attack success rate of the eight original attacks~by~\todo{2.45\%}~at the cost of a slight decrease of the benign accuracy by \todo{0.30\%}. Hence, \tool preserves the capability of existing backdoor attacks. Then,~we measure how \tool affects the survivability of the original attacks when various defenses are applied. On average, \tool significantly enhances the original attacks' survivability from \todo{39.10\%}~to~\todo{62.45\%}.

Finally, we explore the factors that impact the performance~of~\tool; and adopt explainability techniques to delve into the intrinsic~mechanism of \tool, explaining the performance of \tool.

In summary, this work makes the following main contributions.
\begin{itemize}[leftmargin=*]
    \item We have proposed \tool, the first generic approach to enhance the survivability of existing backdoor attacks against existing model reconstruction-based backdoor defenses. 
    \item We have conducted large-scale experiments 
    with eight backdoor attacks and eight backdoor defenses to demonstrate that \tool significantly improves the survivability of attacks against~defenses without impacting the capability of the original attacks.
    \item We have implemented \tool and released the source code of \tool at \url{https://github.com/VenomEnhancer/Venom}.
\end{itemize}

\section{Related Work}

We review the most closely relevant works in two aspects,~i.e.,~backdoor defenses and attacks. We refer readers to recent surveys~\cite{kaviani2021defense, li2022backdoor, Goldblum2023} for a comprehensive discussion about the state-of-the-art.

\subsection{Backdoor Defenses}\label{sec:defenses}

Several backdoor defenses have been proposed to mitigate backdoor threats in DNNs. Mainstream defenses include two categories, i.e., data distribution-based and model reconstruction-based defenses. Data distribution-based defenses are built upon~the assumption that models trained on poisoned datasets tend to learn highly distinct latent representations for poisoned and benign samples in the target class, forming two distinct clusters in various~latent~spaces.~For~example, Tran et al.~\cite{tran2018spectral} and Chen et al.~\cite{chen2018detecting} propose to identify~and~remove poisoned samples via cluster analysis in the latent space, and retrain the model with the remaining samples. Hayase~et~al.~\cite{hayase2021spectre}~and Tang et al.~\cite{tang2021demon} improve the previous two approaches to work~well even at a low poison rate. However, several attacks~\cite{shokri2020bypassing,ijcai2022p0242,xia2022enhancing}  have shown resistance against such data distribution-based defenses.

Model reconstruction-based defenses are usually based on two~assumptions as detailed in Section~\ref{sec:intro}, and use backdoor unlearning~and pruning to mitigate the threats. Given the first observation,~defenders leverage unlearning to correct model's activation behavior.~For example, Liu et al.~\cite{liu2018fine} fine-tune the backdoored model on a subset of benign samples, leveraging the catastrophic forgetting of DNNs to forget the backdoor. Zeng et al.~\cite{zeng2022adversarial} formulate the unlearning~as a mini-max problem, and propose implicit backdoor~adversarial~unlearning to solve the problem. Li et al.~\cite{li2021neural} adopt knowledge distillation to reconstruct the backdoored model. Wang et al.~\cite{wang2019neural} first synthesize the potential trigger and determine the target class, and then exploit this knowledge to unlearn the backdoor.

Given the second observation, defenders identify the neurons or channels in backdoor-related decision paths and prune them.~For~example, Liu et al.~\cite{liu2018fine} propose to prune neurons that are dormant for benign samples because neurons associated with backdoor-related behavior often remain inactive when processing benign samples, and combine pruning with fine-tuning to strengthen~the~defense. Zheng et al.~\cite{zheng2022pre, zheng2022data} observe that the standardized entropy~of~backdoor-related neurons is significantly lower than clean neurons, and channels with high Lipschitz constants is backdoor-related, and hence they prune such neurons or channels to remove the backdoor. Inspired by the mechanism of the optical polarizer, Zhu et al.~\cite{zhu2023neural}~insert a learnable neural polarizer into the backdoored model as an intermediate layer, filtering trigger information and therefore blocking the subsequent backdoor-related decision paths.

There also exist several other defenses. Certified defenses~are~often based on random smoothing techniques~\cite{wang2020certifying}~or~ensemble techniques~\cite{jia2021intrinsic,levine2020deep,jia2020certified} to provide a theoretical guarantee under certain assumptions. However, they are generally~weaker than the above empirical defenses in practice~\cite{li2022backdoor}. Poison suppression-based defenses~\cite{li2021anti,huang2021backdoor} attempt to prevent the creation of hidden backdoors by depressing the effectiveness of poisoned samples during~the~training process. However, they assume that defenders have control over the training process. Preprocessing-based defenses~\cite{8119189,doan2020februus,udeshi2022model,villarreal2020confoc}~use trigger override or data augmentation to prevent the backdoor from being activated in the inference phase. However, they focus on attacks that use local patches as triggers, and thus are less applicable to defend a wide range of attacks.

\subsection{Backdoor Attacks}\label{sec:attacks}

Gu et al.~\cite{gu2019badnets} introduce the first backdoor attack, named {BadNets},~in DNNs. It generates poisoned samples by simply stamping the fixed backdoor trigger onto the selected benign samples and changing their labels to attacker-specified target labels, and hence it is considered as a \textit{poison-label visible attack}. Since then, various similar backdoor attacks have been proposed. For example, Chen et al.~\cite{chen2017targeted} propose to blend instead of stamp the fixed backdoor trigger with benign samples; and Liu et al.~\cite{liu2018trojaning} propose to generate a general trojan trigger by inversing the neural network.

Specifically, one thread of works attempt to achieve \textit{clean-label attacks}, where the poisoned samples are not mislabeled but correctly labeled to evade label inspection. For example, Turner~et~al.~\cite{turner2019label}~propose to poison the benign samples via adversarial perturbations and generative models with the goal of rendering them hard~to~classify and keeping the modification small to ensure the consistent~label. Another thread of works try to realize \textit{sample-specific~attacks},~where different poisoned samples have different backdoor triggers.~For~example, Nguyen and Tran~\cite{nguyen2020input} argue that fixed triggers~are~easily detected, and propose the first sample-specific backdoor attack.~They develop an input-aware trigger generator driven by diversity loss. To further make the sample-specific triggers invisible and improve the stealthiness, Nguyen and Tran~\cite{nguyen2021wanet} use image warping to inject backdoor triggers, and Li et al.~\cite{li2021invisible} use DNN-based image steganography to generate invisible sample-specific backdoor triggers.

With the advancement of backdoor defense techniques, backdoor attacks have started to focus on the resistance against backdoor~defenses.
Several techniques~\cite{shokri2020bypassing,ijcai2022p0242,xia2022enhancing} have been proposed to reduce the latent separation between poisoned and benign samples, thus circumventing existing data distribution-based defenses. However, these attacks assume additional control over the training process. Instead, Qi~et~al.~\cite{qi2022revisiting} reduce the latent separation in poison-only attacks which only assume control over a small part of training~data.

In summary, the existing attacks merely focus on optimized~trigger and data distribution, but barely consider the resistance~to~model reconstruction-based defenses. Our work fills this gap by developing a generic enhancer that not only improves the survivability~of the existing attacks against various model reconstruction-based~defenses but also preserves the existing~attacks' original characteristics (e.g., trigger stealthiness and latent inseparability). 

\section{Threat Model}\label{sec:threat}

\label{sec:threat_model}
\begin{figure}[!t]
    \centering
    \includegraphics[width=.47\textwidth]{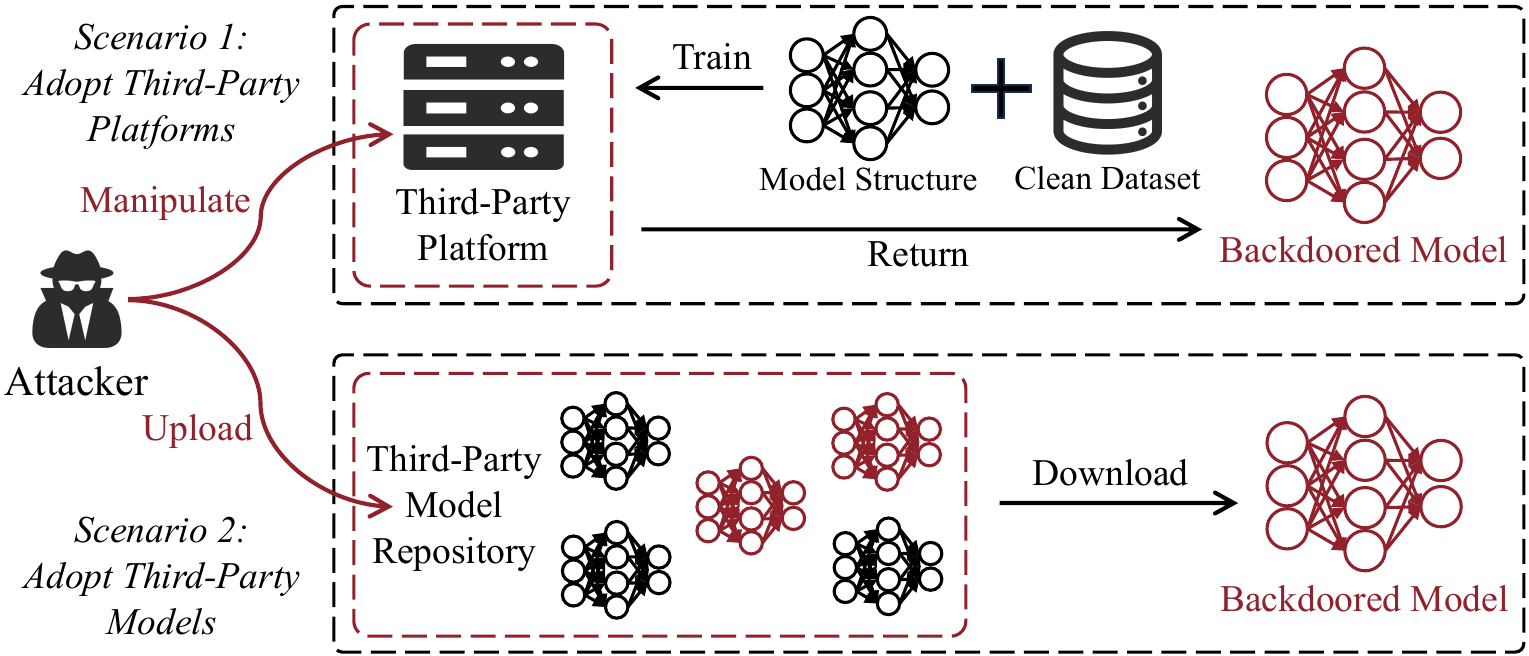}
    \vspace{-5pt}
    \caption{The threat model of \tool on two typical scenarios. During the backdoor injection, the attacker can only control the pipeline within the attacker's capabilities (red areas), but cannot change any victim's behaviors (black areas).}
    \label{fig:threat_model}
\end{figure}

\textbf{Attacker's Goals.} The attacker aims to enhance existing backdoor attacks to craft a backdoored model that achieves three goals.

\begin{itemize}[leftmargin=*]
\item \textit{Effectiveness Goal.} The backdoored model should predict the~poisoned sample with the specific trigger as the target label. 

\item \textit{Utility Goal.} The backdoored model should maintain the same predictive power as a clean model on benign samples.

\item \textit{Survivability Goal.} The backdoored model should survive from existing model reconstruction-based defenses. 
Specifically, after taking defensive measures, the backdoored model should still be able to achieve the effectiveness goal, or at least fail to achieve~the utility goal (i.e., making the backdoored model unusable). 
\end{itemize}

\textbf{Attacker's Capabilities.} As a generic backdoor attack enhancement technique, attacker's capabilities should be applicable to most attacks. We consider the following two typical scenarios where~backdoor~threats~could occur~\cite{li2022backdoor}, as shown in Figure~\ref{fig:threat_model}.

\begin{itemize}[leftmargin=*]
\item \textit{Scenario 1: Adopt Third-Party Platforms.} Users provide their clean dataset, model structure, and training schedule to a potentially untrusted third-party platform (e.g., Google Cloud) to train their model. The attacker can hack into the platform (or the attacker is the service provider of the platform), and can manipulate~the~clean dataset and training schedule to inject the backdoor. However, the attacker cannot change the model structure; otherwise, users will be aware of the change and notice the attack. After training, users will get the backdoored model from the platform.

\item \textit{Scenario 2: Adopt Third-Party Models.} The attacker can upload~the backdoored model to any third-party model repository (e.g., Hugging Face). In that sense, the attacker can have the full access to the whole training process, including the dataset, model structure, and training schedule. 

\end{itemize}

\textbf{Defender's Capabilities.} In both scenarios, the defender cannot control the training process, but can modify the backdoored model. In Scenario 1, the defender can also access the clean dataset.

\section{Methodology}\label{sec:3}

We first formalize backdoor attack enhancer \tool, then introduce the overview of \tool, and finally elaborate each step of \tool.

\subsection{Formalization of \tool}\label{sec:formal}

We first formalize \tool as a bi-level optimization problem,~and~then formalize \tool as a binary-task optimization problem.

\textbf{Bi-Level Optimization Problem}. In classic DNN training,~a~clean model $f:\mathcal{X} \rightarrow \mathcal{Y}$ is trained on a clean dataset $\mathcal{D}_c$ to fit the~data~distribution $\mathcal{X} \times \mathcal{Y}$, where $\mathcal{X}$ denotes the input space, $\mathcal{Y}$ denotes~the~output space representing class labels, and $\mathcal{D}_c$ denotes a set of data~samples $(x,y)\in \mathcal{X} \times \mathcal{Y}$. To achieve the attacker's goals,~the~attacker~needs to inject a backdoor into the clean model $f$ to generate~a~backdoored model $f'$ which is trained on a backdoored dataset $\mathcal{D}_{bd}$.~The backdoored dataset $\mathcal{D}_{bd}$ is generated by poisoning a certain rate~of data samples in the clean dataset $\mathcal{D}_c$. The data poisoning maps samples with the backdoor trigger $\tau$ to the target label $t$, with~the~poisoned sample generator $\mathcal{G}:\mathcal{X}\rightarrow \mathcal{X}$ and the label flipper $\mathcal{F}:\mathcal{X} \times \mathcal{Y} \rightarrow \mathcal{Y}$. $\mathcal{J} = \{j_1, \cdots, j_k\} $ denotes the indices of the $k$ poisoned samples,~and these poisoned samples are denoted as $\mathcal{D}_{bd}^p$. The remaining~samples are benign samples, denoted as $\mathcal{D}_{bd}^b$. Therefore, $\mathcal{D}_{bd} = \mathcal{D}_{bd}^p \cup \mathcal{D}_{bd}^b$, and each data sample $(x_i',y_i')\in \mathcal{D}_{bd}$ is formulated by Eq.~\ref{eq:sample}.
\begin{equation}\label{eq:sample}
    \begin{aligned}
        x_i'=\begin{cases}
            \mathcal{G}(x_i), \ i\in \mathcal{J}  \\
            x_i,\ otherwise
            \end{cases},\ \ \ \ 
        y_i'=\begin{cases}
            \mathcal{F}(x_i,y_i), \ i\in \mathcal{J}  \\
            y_i,\ \ \ \ \ otherwise
            \end{cases}
    \end{aligned}
\end{equation}

We design \tool as a generic backdoor attack enhancer,~and~formalize it as a bi-level optimization problem, as formulated by Eq.~\ref{eq:bi-optimization},
\begin{equation}
    \label{eq:bi-optimization}
    \begin{aligned}
        \min 
        \ \ \ \ \ \ \ \ 
        \mathcal{L}_{out}\left(\mathcal{D}_{bd},f^{'*}\right) &= \sum_{(x',y')}^{\mathcal{D}_{bd}} \mathbb{I}(f^{'*}(x')\ne y'), \\
        \mathrm{s.t.} 
        \ \ \ \ \ \ \ \ \ \ \ \ \ \ 
        f^{'*} \in \mathop{\arg\min}_{f'}& \ \mathcal{L}_{inner}\left(\mathcal{D}_{bd}[t],f,f'\right),\\
        \mathrm{where} 
        \ \ \ \ \ \ \ \ \ \ \ \ \ \ \ \ 
        \mathcal{L}_{inner} = &\sum_{(x',y')}^{\mathcal{D}_{bd}[t]}\mathcal{S}\left(x',x_{ref},f',f\right)\\
    \end{aligned}
\end{equation}
where $\mathbb{I}(\cdot)$ denotes the indicator function, i.e., $\mathbb{I}(\cdot) = 1$ if $\cdot$ is true~and $\mathbb{I}(\cdot) = 0$ if $\cdot$ is false; $\mathcal{D}_{bd}[t]$ denotes the set of samples with~the~target label $t$ in $\mathcal{D}_{bd}$, including both benign and poisoned samples; $x_{ref}$ denotes a reference sample with the target label $t$, which~is~randomly selected from $\mathcal{D}_{c}$; and $\mathcal{S}$ denotes~a~similarity function. 

In outer optimization, we minimize the loss function $\mathcal{L}_{out}$, which measures the number of samples that are not predicted as the~corresponding label $y'$ by the optimal backdoored model $f^{'*}$. Specifically, $\mathcal{L}_{out}$ not only forces $f^{'*}$ to predict the correct label $y'$ for each benign sample, which satisfies the attacker's utility goal,~but~also~forces $f^{'*}$ to identify the trigger $\tau$ and predict the target label $t$ for each~poisoned sample, which satisfies the attacker's effectiveness goal.

In inner optimization, we minimize the loss function $\mathcal{L}_{inner}$~to~find the optimal backdoored model $f^{'*}$ which satisfies the attacker's~survivability goal. To this end, we force the decision paths and activation values of both~poisoned and benign samples in $\mathcal{D}_{bd}[t]$~in~the~backdoored model~$f'$~to be similar to those of reference samples~in~the~clean model $f$.~Hence, poisoned samples and benign samples in $\mathcal{D}_{bd}[t]$ have similar decision paths and activation values in $f'$, which breaks the assumptions of existing model reconstruction-based defenses and thus makes $f'$ survive from those defenses.

\textbf{Binary-Task Optimization Problem.} It is difficult~to~guarantee the optimal solution for the bi-level optimization problem,~and~only strong stationary solution can be obtained~\cite{liu2022investigating}. Moreover, the solving process is complicated and requires prohibitive time overhead. Therefore, we relax it into a binary-task optimization problem~to find an approximate solution, using the joint loss function in Eq.~\ref{eq:total_loss},
\begin{equation}
    \label{eq:total_loss}
    \mathcal{L} = \omega_1 \cdot \mathcal{L}_1 + \omega_2 \cdot \mathcal{L}_2
\end{equation}
where $\mathcal{L}_1$ is the loss function of the first task, i.e., the \textit{original~backdoor attack task}, $\mathcal{L}_2$ is the loss function of the second task, i.e.,~the~\textit{attack enhancement task}, and $\omega_1$ and $\omega_2$ are the weight to each task. 

In the first original~backdoor attack task, \tool aims to preserve the original attack capability, i.e., to inject a backdoor~and~satisfy~the effectiveness and utility goals. This corresponds to the outer optimization in bi-level optimization. Existing backdoor attacks can be unified into the minimization of the loss function $\mathcal{L}_1$ in Eq.~\ref{eq:l1},
\begin{equation}
    \label{eq:l1}
    \mathcal{L}_1 =  \sum_{(x',y')}^{\mathcal{D}_{bd}} \mathop{l}\left(x',y',f'\right) + l_{tc}
\end{equation}
where the loss $\mathop{l}(x', y', f')$ measures the difference between~the~backdoored model's predicted label $f'(x')$ and the ground truth label $y'$. In multi-classification, the cross-entropy loss is commonly adopted as $l$.
Apart from data poisoning, some backdoor attacks also control the training phase and introduce the additional loss $l_{tc}$, which we also inherit to keep the capability of original backdoor attacks.

\begin{figure*}[!t]
    \centering
    \includegraphics[width=.85\textwidth]{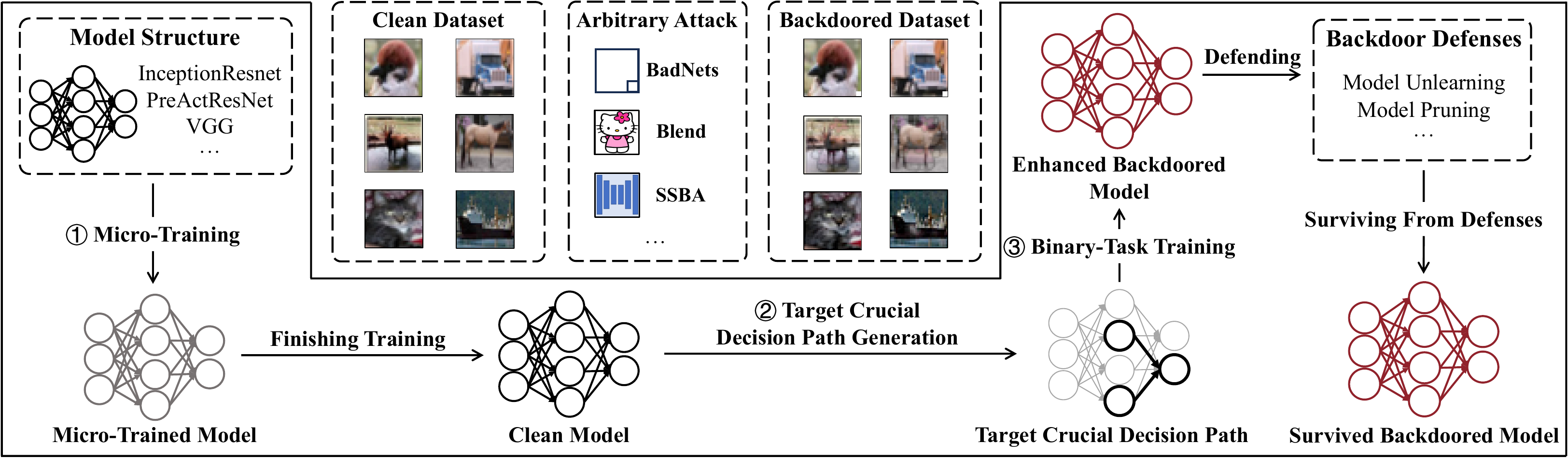}
    \vspace{-5pt}
    \caption{Approach overview of \tool.}
    \label{fig:Overview}
\end{figure*}

In the second attack enhancement task, \tool aims to enhance the attack survivability from model reconstruction-based defenses, i.e., to satisfy the survivability goal. This corresponds to the inner optimization in bi-level optimization. The loss function $\mathcal{L}_2$ of this task will be introduced in Section~\ref{sec:3.4}.

\subsection{Overview of \tool}

Based on the formalization of the binary-task optimization,~we~present the approach overview of \tool in Figure~\ref{fig:Overview}. Specifically, \tool~first conducts \textit{micro-training}, i.e., training the model on the clean~dataset for a few epochs. The resulting micro-trained model will be used~to initialize the model in backdoor injection and enhancement. Then,~it finishes the training on the clean dataset to obtain a clean model,~and conducts \textit{target crucial decision path generation} to produce the decision path (i.e., a set~of neurons) that is crucial for classifying benign samples with the target label from a clean model. Finally,~it~conducts \textit{binary-task training} to inject backdoor and enhance attack~survivability. For the original backdoor attack task, it uses~the original~loss, as formulated in Section~\ref{sec:formal}. For the attack enhancement~task,~it~uses a novel \textit{attention imitation loss} to force the decision path of poisoned samples in the backdoored model to couple with the generated crucial decision path of benign samples with the target label.~This~coupling ensures that the behavior of our enhanced backdoored model is closely similar to that of the clean model, thereby enabling it to survive from existing model reconstruction-based defenses.~In~the following sections, we present the three key steps: micro-training, target crucial decision path generation, and binary-task training.

\subsection{Micro-Training}\label{sec:micro}

In the micro-training step, we train the model on~the clean~dataset for a few epochs (e.g., 5\% of the whole epochs). This resulting early-stage clean model, referred to as a micro-trained~model $f_m$, already exhibits some level of performance, but has not fully learned the features from the clean dataset. Instead of training the enhanced backdoored model from scratch in the later binary-task training~step (see Section~\ref{sec:3.4}), we utilize this micro-trained model as the initialization.
This design is inspired by task scheduling~in~multi-task~learning~\cite{jean2019adaptive}. 
Micro-training helps \tool to establish basic model classification capability and feature representations before introducing the backdoor attack task and attack enhancement task.

Specifically, micro-training mainly plays two roles. First,~it~guides the overall direction of our binary-task optimization. The two tasks exhibit significant differences in complexity, with the more complex backdoor attack task dominating the optimization direction.~If~the model is initialized~randomly, this dominance results in premature overfitting to the backdoor attack task, while the attack enhancement task fails to be effectively trained. Our micro-training~reduces the learning complexity of the backdoor attack task, as it shares similarities with the learning objective of the backdoor attack task. Consequently, it to some extent balances the complexity and learning speed of the two tasks. In this way, the overall optimization~direction does not excessively favor the backdoor attack task, preventing the model from getting stuck in local optima, and instead follows a more balanced direction. Second, it facilitates the inclination of the enhanced backdoored model in binary-task optimization towards the clean model's behavior, which also contributes to enhancing the model's survivability to some extent.

\subsection{Target Crucial Decision Path Generation}\label{sec:3.3}

\begin{algorithm}[!t]
    \caption{Target Crucial Decision Path Generation}
    \label{alg:1}
    \begin{algorithmic}[1]
    \small
        \Require clean model $f$, neurons in a selected layer $\mathcal{N}_{layer}$,
        clean dataset $\mathcal{D}_{c}$, target class $t$, 
        total number of neurons in the TCDP $k$
        \Ensure a set of neurons that form the TCDP $\mathcal{N}$
        \LComment{Select crucial neurons for each class} 
        \For {each class $i$} \label{alg:1:foreachclass}
        \State $\mathcal{S}_{i} \gets $ \Call{CalcSimilarity}{$f, \mathcal{N}_{layer},\mathcal{D}_{c}[i]$} \label{alg:1:foreachclass:in}
        \State $\mathcal{N}_{i} \gets \{n \mid \mathcal{\mathcal{S}}_{i}[n] > \epsilon_1\}$
        \EndFor \label{alg:1:foreachclass:end}
        \LComment{Remove common neurons} 
        \State $S_t \gets$ \Call{CalcSimilarity}{$f, \mathcal{N}_{t}, \mathcal{D}_{c}$} \label{alg:1:remove1}
        \State $\mathcal{N}_{t} \gets \{n \mid n \in \mathcal{N}_{t}$ and $ \mathcal{S}_t[n] < \epsilon_2 \}$ \label{alg:1:remove1:end}
        \LComment{Select neurons that form the TCDP}
        \For {each neuron $n \in \mathcal{N}_{t}$}               \label{alg:1:count}
        \State $Count[n] \gets$ the number of classes that consider $n$ as crucial
        \EndFor
        \State $\mathcal{N} \gets$ top $k$ neurons in $\mathcal{N}_{t}$ sorted by $Count$      \label{alg:1:count:end}
        \State \Return $\mathcal{N}$ 
        \Function{CalcSimilarity}{$f, \mathcal{N}_{candidate}, \mathcal{D}$} \label{alg:1:calcsimilarity}
        \State $\mathcal{S} \gets \emptyset$
        \State $\mathcal{D}^a,\mathcal{D}^b \gets$ divide $\mathcal{D}$ into two groups randomly \label{alg:1:divide}
        \For {each neuron $n \in \mathcal{N}_{candidate}$}
            \State $s \gets 0$
            \LComment{Calculate the similarity of each sample pair}
            \For {each sample pair $(x_a,x_b) \in \mathcal{D}^a \times \mathcal{D}^b$} \label{alg:1:sim1}
                \State $s \gets s + \mathop{sim}(\sigma_n(x_a,f),\sigma_n(x_b,f))$ \label{alg:1:sim2}
            \EndFor
            \State $\mathcal{S}[n] \gets s / (\mathop{size}(\mathcal{D}^a) *  \mathop{size}(\mathcal{D}^b))$ \label{alg:1:sim3}
        \EndFor
        \State \Return $\mathcal{S}$
        \EndFunction \label{alg:1:calcsimilarity:end}
    \end{algorithmic}
\end{algorithm}

Given a clean model, this step aims to generate the decision path~that is crucial~for~classifying benign samples with the target label. Such~a decision path is referred to as target crucial decision path (TCDP). Typically, decision path is composed of neurons (e.g.,~convolutional kernels in a CNN). According to the literature on DNN~\cite{khan2020survey},~the~shallow layers of the network extract abstract features from the samples, while the deep layers are capable of extracting specific~features~from the samples. Therefore, we select neurons in a single deep layer as candidate neurons, from which we further calculate~the~TCDP.

After determining the single deep layer, we select a fixed number of neurons to form the TCDP. We select neurons that exhibit similar activation values for benign samples with the target label, meaning that they are crucial for the classification of the target~class.~Further, we exclude common neurons which have similar activation~values across all sample classes and thus~are~not~crucial~for~classification. We also favor neurons~that~also~play~important roles in classification of some other classes. Our selection~aims~to couple the decision~path of poisoned samples with that of benign samples from multiple classes in binary-task training (see Section~\ref{sec:3.4}), thereby~increasing the difficulty and cost of implementing effective defenses.

Algorithm~\ref{alg:1} shows the procedure to generate the TCDP. First,~for each class $i$, it computes the similarity of activation values of samples in this class $\mathcal{D}_c[i]$ on all neurons $\mathcal{N}_{layer}$ in a selected~layer~(Line \ref{alg:1:foreachclass}-\ref{alg:1:foreachclass:in}) through the \textsc{CalcSimilarity} function (Line~\ref{alg:1:calcsimilarity}-\ref{alg:1:calcsimilarity:end}). Specifically, this function first randomly divides the dataset $\mathcal{D}$ into two~groups $\mathcal{D}^a$ and $\mathcal{D}^b$ (Line~\ref{alg:1:divide}), then computes the cosine similarity $sim(\cdot, \cdot)$~of the activation values $\sigma_n(\cdot, \cdot)$ of each sample pair $(x_a,x_b)$ on each candidate neuron $n$ (Line~\ref{alg:1:sim1}-\ref{alg:1:sim2}), and finally calculates the average activation value similarity of each sample pair on each neuron~(Line \ref{alg:1:sim3}). Next, for each class $i$, Algorithm~\ref{alg:1} proceeds to select crucial neurons $\mathcal{N}_{i}$ (Line~\ref{alg:1:foreachclass:end}), which consists of neurons exhibiting an average activation value similarity greater than the threshold $\epsilon_1$. 

Next, Algorithm~\ref{alg:1} computes the similarity of activation values of the samples from the entire clean dataset $\mathcal{D}_c$ on the crucial neurons of the target class $\mathcal{N}_{t}$ (Line~\ref{alg:1:remove1}), and removes the common neurons whose activation value similarity across the entire clean dataset exceeds the threshold $\epsilon_2$ (Line~\ref{alg:1:remove1:end}). In other words, the removed~common neurons exhibit similar activation behavior~for~all~classes.

Finally, Algorithm~\ref{alg:1} sorts each crucial neuron $n$ in $\mathcal{N}_{t}$ based~on the number of classes whose crucial neurons also contain $n$, meaning that $n$ is also important for some other classes, and selects the top $k$ neurons $\mathcal{N}$ to form the TCDP (Line~\ref{alg:1:count}-\ref{alg:1:count:end}).

\subsection{Binary-Task Training}\label{sec:3.4}

The binary-task training step involves two tasks. The first task~is~to inject backdoor, which uses the loss of the original backdoor attack as formulated in Eq.~\ref{eq:l1}. The second task is to enhance attack~survivability, which uses our novel attention imitation loss. We~also~optimize the binary-task training to better balance the two tasks. 

\begin{figure}[!t]
    \centering
    \includegraphics[width=.43\textwidth]{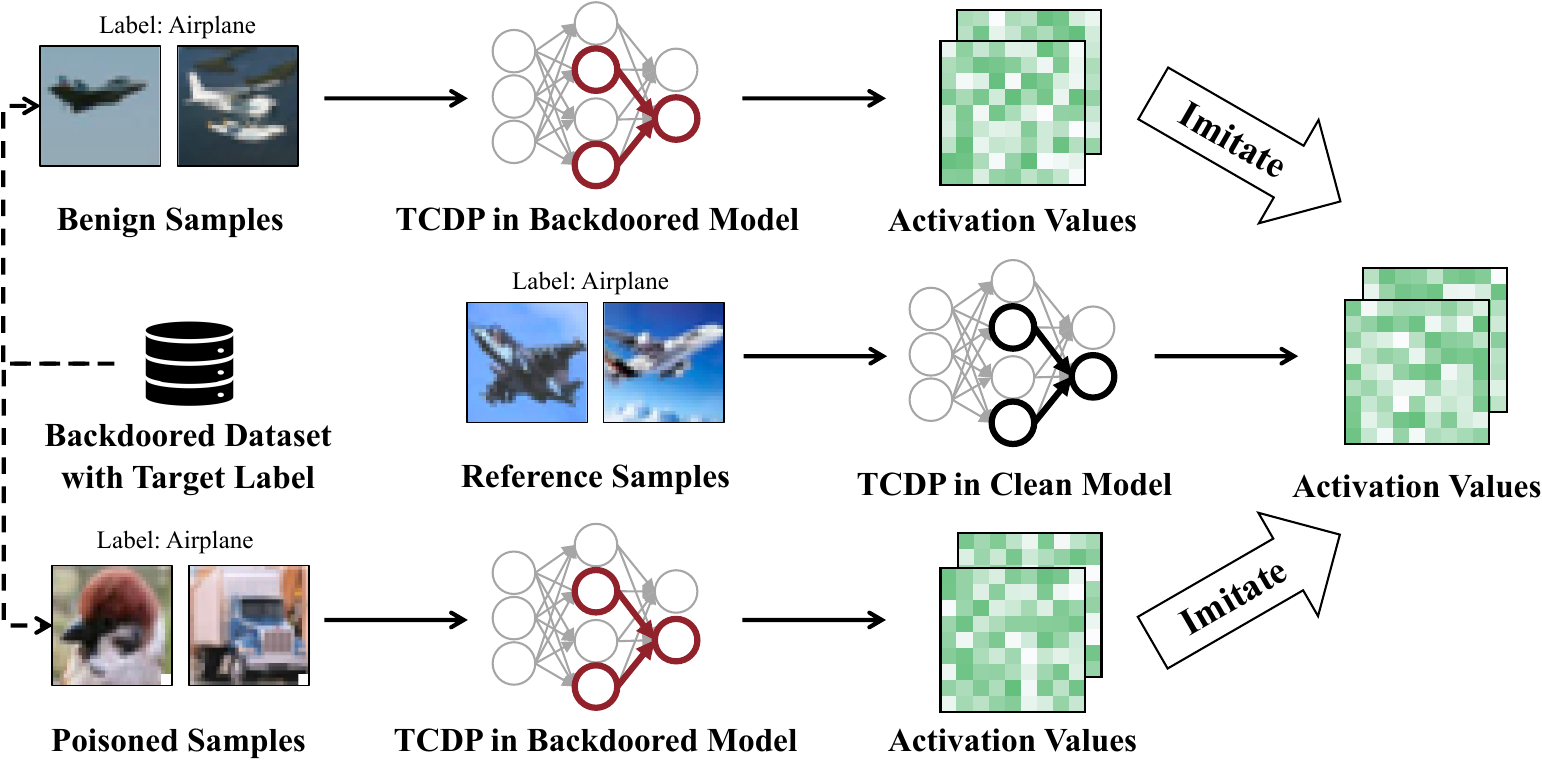}
    \vspace{-5pt}
    \caption{Illustration of attention imitation loss.}
    \label{fig:AIL}
\end{figure}

\textbf{Attention Imitation Loss.} To make a backdoor attack survive from defenses, our goal is to force the behavior of poisoned~samples in the backdoored model to imitate that of benign samples.~To~achieve this goal, we introduce a novel \textit{attention imitation loss}. It is inspired by attention transfer loss~\cite{zagoruyko2017paying}, but involves two adaptations. First, we refine the computation granularity to the level of neurons instead of the layer level in attention transfer loss,~which~enables~more precise control over activation values. Besides, attention imitation loss is only calculated on neurons in the TCDP, reducing the impact on the backdoor attack task. Second, attention transfer loss~transfers the attention from one model to another for~the~same sample, while we achieve attention imitation across different samples. Specifically, the activation behavior of poisoned samples imitates that of benign samples with the target label. Therefore, it helps to evade defenses.

Formally, we define attention imitation loss ($\mathcal{AIL}$) by Eq.~\ref{eq:l2},
\begin{equation}
    \label{eq:l2}
    \begin{aligned}
        \mathcal{AIL} &=  \sum_{n}^{\mathcal{N}} \parallel \frac{\sigma_n(x',f')}{\parallel \sigma_n(x',f') \parallel_2} -  \frac{\sigma_n(x_{ref},f)}{\parallel \sigma_n(x_{ref},f) \parallel_2} \parallel_1\\
        \mathcal{L}_2 &= \sum_{(x',y')}^{\mathcal{D}_{bd}[t]} \mathcal{AIL}(x',x_{ref},f',f)
    \end{aligned}
\end{equation}
where $\sigma_n(\cdot, \cdot)$ denotes the activation value of the neuron $n$; $\mathcal{D}_{bd}[t]$ denotes the set of samples with~the~target label $t$ in the backdoored dataset $\mathcal{D}_{bd}$; $x'$ denotes a sample in $\mathcal{D}_{bd}[t]$; $x_{ref}$ denotes a reference sample with the target label $t$, which is randomly selected~from the clean dataset $\mathcal{D}_{c}$; and $f$ and $f'$ denote the clean model and the backdoored model, respectively.

For each neuron $n$ in the TCDP $\mathcal{N}$ (see Section~\ref{sec:3.3}), $\mathcal{AIL}$ first calculates the $l_2$ normalization of the activation values of $x'$ in $f'$ and $x_{ref}$ in $f$ on neuron $n$, respectively. Then, $\mathcal{AIL}$ computes~the $l_1$ normalization of their difference. Based on $\mathcal{AIL}$, the loss function $\mathcal{L}_2$ of the attack enhancement task is computed over each~sample in $\mathcal{D}_{bd}[t]$. As illustrated in Figure~\ref{fig:AIL}, $\mathcal{AIL}$ plays two roles, one~is~for benign samples, and the other is for poisoned samples.

If $x' \in \mathcal{D}_{bd}^b$ (i.e., $x'$ is a benign sample), $\mathcal{AIL}$ guides its behavior in the backdoored model towards that in a clean model, as illustrated in the upper part of Figure~\ref{fig:AIL}. Specifically, due~to~the~injection of the backdoor, the decision paths of $x'$ in $f'$ are prone~to~deviate. This deviation can be exploited for backdoor detection. $\mathcal{AIL}$ assists in rectifying the deviated paths in $f'$ to achieve survivability.

\begin{figure}[!t]
    \centering
    \includegraphics[width=.43\textwidth]{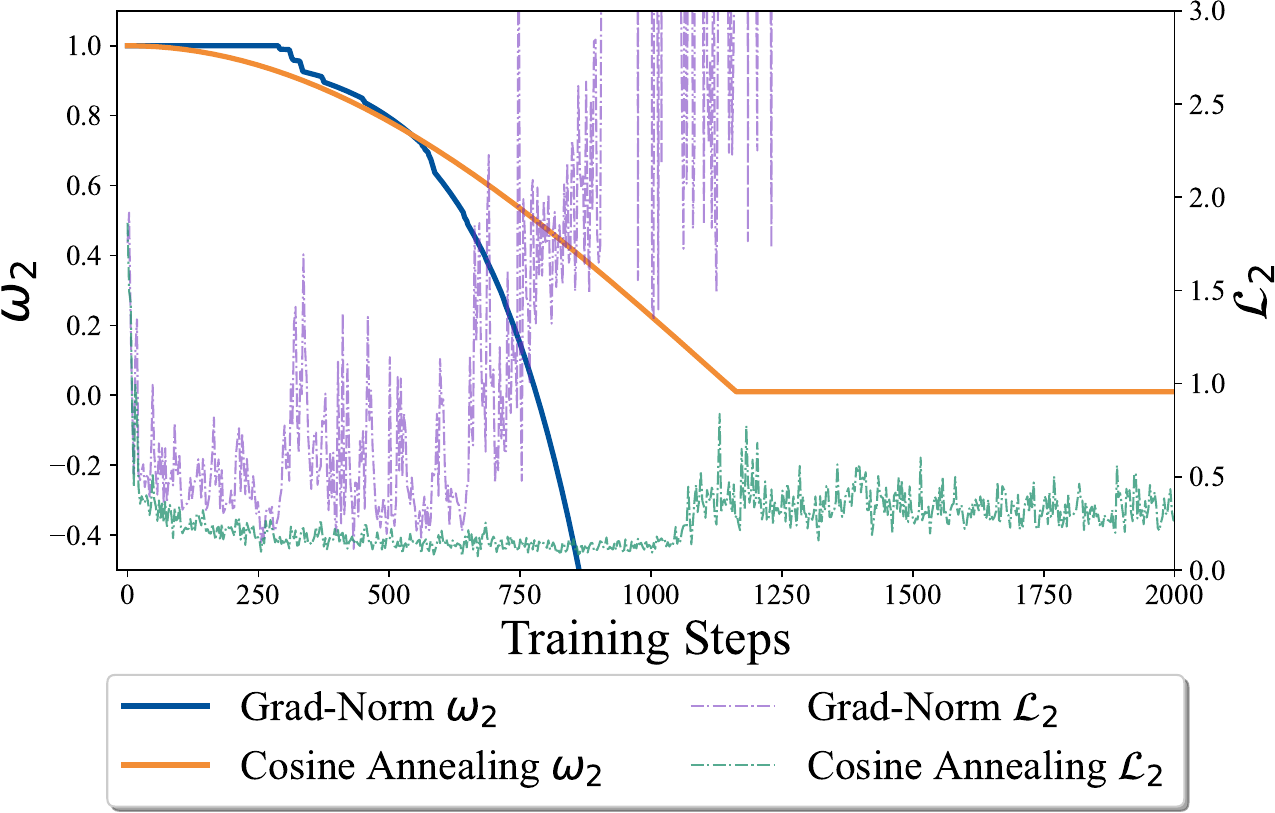}
    \vspace{-5pt}
    \caption{Training process using different strategy.}
    \label{fig:cos}
\end{figure}

If $x' \in \mathcal{D}_{bd}^p$ (i.e., $x'$ is a benign sample), $\mathcal{AIL}$ forces poisoned samples and benign samples to behavior similarly~on~the~TCDP~$\mathcal{N}$,~as illustrated in the bottom part of Figure~\ref{fig:AIL}. This brings two distinct~advantages. One advantage is to evade backdoor detection techniques such as gradient and activation map analysis, since benign samples and poisoned samples show similar behavior in $f'$. The other~advantage is the tight coupling between the backdoor and the target~class. The intuition behind is that $\mathcal{AIL}$ guides the decision~path~of~poisoned samples partially overlaps with that of benign samples in the target class, thereby increasing the coupling degree between the backdoor and the classification accuracy of the target class.~Additionally, the neurons in the TCDP $\mathcal{N}$ are not only crucial for~the~classification of the target class, but part of them are also crucial~for~the classification of other classes (see Section~\ref{sec:3.3}),~making the backdoor further coupled with the overall classification accuracy. Therefore, the backdoor becomes difficult~for~defense~techniques to eliminate. Even if certain defense technique successfully eliminates~the~backdoor, it will inevitably lead to a significant decrease in the classification accuracy of benign samples. Ultimately, this renders defense techniques ineffective in mitigating the backdoor.

\textbf{Optimized Binary-Task Training.} We use the joint loss function in Eq.~\ref{eq:total_loss} to further train the micro-trained model $f_{m}$ (see Section~\ref{sec:micro}) into the enhanced backdoored model $f'$.

As shown by Eq.~\ref{eq:total_loss}, we follow the common practice of multi-task learning to assign a weight $\omega_i$ to the loss of each task. By employing this approach, we are able to adjust the significance for each task manually. However, the fixed value of $\omega_i$ persists throughout the entire training process, which also has its limitations. Given that the two tasks can vary in their learning difficulty, they may exhibit differences in convergence rates and magnitudes. Specifically, as $\mathcal{L}_2$ is computed on a limited set of neurons, $\mathcal{L}_2$ exhibits a smaller scale and faster convergence rate compared to $\mathcal{L}_1$. If $\mathcal{L}_2$ is close to convergence but $\mathcal{L}_1$ has not yet, the subsequent training process will shift towards the optimization direction of  $\mathcal{L}_1$.
Consequently, $f'$ will only find the local optimum, impacting the effectiveness of the $\mathcal{L}_2$ task.
Therefore, the weight $\omega_i$ should be dynamic.

Previous work~\cite{chen2018gradnorm} adopts gradient normalization (Grad-Norm)~to balance multiple tasks, which places gradient norms for different tasks on a common scale and dynamically adjusts gradient norms~so that different tasks train at a similar rate. However, this strategy is not suitable for our training scenario as it may lead to an unstable training process and hinder model convergence.
As shown in Figure~\ref{fig:cos}, due to the extreme imbalance between the two tasks, $\omega_2$ drops sharply and even becomes negative, while~$\mathcal{L}_2$~fluctuates~violently and quickly exceeds the calculable range.

\begin{algorithm}[!t]
    \caption{Optimized Binary-Task Training}
    \label{alg:2}
    \begin{algorithmic}[1]
    \small
        \Require  micro-trained model $f_{m}$, clean model $f$, clean dataset $\mathcal{D}_{c}$,
        backdoored dataset $\mathcal{D}_{bd}$, 
        target class $t$,
        neurons in the TCDP $\mathcal{N}$
        \Ensure enhanced backdoored model $f'$   
        \State Initialize the enhanced backdoored model $f'$ as $f_{m}$ \label{alg:2:init}
        \State $\omega_1 \gets 1$, $\omega_2 \gets 1$ \label{alg:2:init:end}
        \State $T \gets s \cdot \frac{\left | \mathcal{D}_{bd} \right |}{batch\_size}$ \label{alg:2:init:2}
        \For{$t$ = 0 to $total\_training\_steps$} \label{alg:2:loopstart}
            \State Input batch $x_i$ to compute $\mathcal{L}_1(t)$ and $\mathcal{L}_2(t)$ \label{alg:2:compute}
            \LComment{Adjust losses to a similar scale}
            \If{$t = 0$} \label{alg:2:lambda}
            \State $\beta \gets \frac{\mathcal{L}_1(0)}{\mathcal{L}_2(0)}$ \label{alg:2:beta1}
            \EndIf \label{alg:2:lambda:end}
            \State $\mathcal{L}_2(t) = \beta \cdot \mathcal{L}_2(t)$ \label{alg:2:beta}
            \LComment{Use cosine annealing strategy to adjust weights}
            \If{$\omega_2 > 0.01$} \label{alg:2:cos}
            \State $\omega_2 \gets \cos (\frac{\pi }{2}\cdot\frac{t}{T})$ \label{alg:2:cos:end1}
            \State $\omega_1 \gets 2 - \omega_2$ \label{alg:2:cos:end2}
            \EndIf
            \State $\mathcal{L}(t) \gets \omega_1 \cdot \mathcal{L}_1(t) + \omega_2 \cdot \mathcal{L}_2(t)$
            \State Update the network $f'$ to minimize $\mathcal{L}(t)$ \label{alg:2:loopend}
        \EndFor
    \end{algorithmic}
\end{algorithm}

Instead, we design a smoother cosine annealing strategy~to~achieve a better balance, which is surprisingly effective despite its simplicity, as illustrated in Figure~\ref{fig:cos}. Algorithm~\ref{alg:2} presents the detailed~procedure of our optimized binary-task training. It first initializes~the~enhanced backdoored model $f'$ as the micro-trained model $f_m$,~and~initializes $\omega_1$ and $\omega_2$ as 1 (Line~\ref{alg:2:init}-\ref{alg:2:init:end}). Then, it sets the maximum training steps $T$ to adjust the weights (Line~\ref{alg:2:init:2}). 
The number of training steps for one epoch is obtained by dividing the size of the backdoored dataset $\mathcal{D}_{bd}$ by the batch size $batch\_size$, and $T$ is the sum of this training steps for $s$ epochs. This means that we adjust the weights of the two tasks within $s$ epochs, which is similar to the effect of Grad-Norm~\cite{chen2018gradnorm}. Here, $s$ is determined by the imbalance between~the two tasks. The greater the imbalance, the smaller the value of $s$.

Next, Algorithm~\ref{alg:2} updates the network $f'$ to inject backdoor and enhance attack survivability by multiple steps (Line~\ref{alg:2:loopstart}-\ref{alg:2:loopend}).~At~each step $t$, it first computes the loss of the two tasks, $\mathcal{L}_1(t)$ and $\mathcal{L}_2(t)$ (Line~\ref{alg:2:compute}). Then, it balances the overall loss magnitudes by $\beta$ (Line~\ref{alg:2:beta}), otherwise the backdoor attack task will overwhelm the other task during the training~\cite{vandenhende2021multi}. The scaler $\beta$ is computed in the first step (Line~\ref{alg:2:beta1}).
In each subsequent step, $\mathcal{L}_2(t)$ is scaled by multiplying it with $\beta$ (Line~\ref{alg:2:beta}). Next, it decays the weight $\omega_2$ with a cosine~annealing for each batch until $\omega_2$ decreases to 0.01, and thereafter keeps it constant (Line~\ref{alg:2:cos}-\ref{alg:2:cos:end1}). It also normalizes the two weights, ensuring that they add up to 2 (Line~\ref{alg:2:cos:end2}). Notice that this algorithm is suitable for scenarios where the two tasks are unbalanced.

\section{Evaluation}

To evaluate the performance of \tool, we design our evaluation~to answer the following four research questions.

\begin{itemize}[leftmargin=*]
    \item \textbf{RQ1 Capability Preservation Evaluation:} How does \tool preserve the capability of the original backdoor attacks?
    \item \textbf{RQ2 Survivability Evaluation:} How does \tool enhance the survivability against model reconstruction-based defenses?
    \item \textbf{RQ3 Ablation Study:} What factors influence the performance of \tool?
    \item \textbf{RQ4 Explainability Analysis:} What guarantees the performance of \tool?
\end{itemize}

\subsection{Evaluation Setup}

We evaluate \tool with two widely used DNN structures~and~three popular datasets. In our evaluation, we use \tool to enhance eight backdoor attacks against eight backdoor defenses.

\textbf{Models and Datasets.} We select two DNNs (i.e., VGG19-BN~\cite{simonyan2015very} and PreActResNet18~\cite{he2016identity}) and three datasets (i.e., CIFAR-10 \cite{krizhevsky2009learning},~CIFAR-100~\cite{krizhevsky2009learning} and GTSRB~\cite{6706807}), which are widely used~in~backdoor~literature. We use VGG19-BN for the three datasets, and PreActResNet18 for the CIFAR-10 dataset, achieving image classification tasks. Table \ref{tbl:Datasets} and \ref{tbl:Models} in Appendix~\ref{Appendix:Datasets and Models} report details about datasets~and~models.

\textbf{Attacks and Defenses.} We use two criteria for selecting backdoor attack and defense techniques. First, it should be classic (e.g.,~the pioneer work) or advanced techniques (e.g., recently published work at top venues). Second, it should cover as many categories~as~possible within the taxonomy \cite{li2022backdoor}, aiming to demonstrate the effectiveness of \tool against various types of attacks and defenses.

\begin{itemize}[leftmargin=*]
\item \textit{Backdoor Attack Selection.} We select eight backdoor attacks. Among them, BadNets~\cite{gu2019badnets}, Blend~\cite{chen2017targeted} and TrojanNN~\cite{liu2018trojaning}~are~classic~poison-label visible attacks.~LC~\cite{turner2019label} is a clean-label attack. 
SSBA~\cite{li2021invisible}~is~a poison-based sample-specific invisible attack,  
while Inputaware~\cite{nguyen2020input} and WaNet~\cite{nguyen2021wanet} are training-controlled sample-specific attacks. Adap-Blend~\cite{qi2022revisiting} is a recent attack which improves latent inseparability to bypass data distribution-based defenses. The general idea of each attack is introduced in Section~\ref{sec:attacks}.

\item \textit{Backdoor Defense Selection.} We select eight model reconstruction-based backdoor defenses. Among them, FT~\cite{liu2018fine}, NAD~\cite{li2021neural} and I-BAU~\cite{zeng2022adversarial} are model~unlearning-based defenses. 
NC~\cite{wang2019neural}~is~a~trigger synthesis-based defense, which first synthesizes the trigger and then unlearns the backdoor. 
BNP~\cite{zheng2022pre}, FP~\cite{liu2018fine} and CLP~\cite{zheng2022data} are model pruning-based defenses.  NPD~\cite{zhu2023neural} attempts to block backdoor-related decision paths. 
The general idea of each defense is presented in Section~\ref{sec:defenses}.
\end{itemize}

\textbf{\tool Setup.} For all image classification tasks, we follow~the settings of the original attacks and defenses, and set the poison rate to 0.1 except for Adap-Blend where the poison rate is set~to~0.003. During the training process, we use the SGD optimizer~with~a~batch size of 128, a momentum of 0.9, and a weight decay of 5e-4.~We~set~the initial learning rate to 0.01, and adopt a cosine annealing to set the learning rate for each training step during the whole epochs. The whole epochs are 80 on CIFAR-10 and CIFAR-100, and 60 on GTSRB.

For specific settings of \tool, the micro-training step takes 5\% of the whole epochs. In the TCDP generation step, we choose the last convolutional layer~as~the~target layer, from which we select a total of 10 crucial neurons to form the TCDP; and we set the two thresholds $\epsilon_1$ and $\epsilon_2$ to 0.9 and 0.7. In the binary-task training step, we set $s$ to 3 epochs. More details can be found in Appendix~\ref{Appendix:Attack Setting}.

\textbf{Evaluation Metrics.} As widely used in previous works, we also adopt \textit{Benign Accuracy (BA)} and \textit{Attack Success Rate (ASR)}~to~evaluate the utility and effectiveness of \tool. In addition, we further propose \textit{Attack Survivability Rate (ASuR)} to measure the impact of \tool on enhancing attacks against defenses.

\begin{itemize}[leftmargin=*]

\item \textit{Benign Accuracy (BA)}. BA is the accuracy of the backdoored~model on the benign testing dataset. It is used to evaluate the utility~of~a backdoor attack. Besides, we use \textit{Benign Accuracy Drop (BAD)} to measure the drop in BA of the backdoored model compared~to~the accuracy of the clean model. A lower BAD indicates that~the~backdoor attack has a lower impact on the model's utility.

\item \textit{Attack Success Rate (ASR)}. ASR is the ratio of the number~of~poisoned samples that are successfully attacked to the total number of poisoned samples in the testing dataset. It is used to measure the effectiveness of a backdoor attack.

\item \textit{Attack Survivability Rate (ASuR)}. ASuR is a comprehensive metric that combines ASR and BA to assess the survivability~of~an~attack against a defense. Existing defenses inevitably sacrifice a certain degree of BA while eliminating the backdoor~\cite{liu2018fine}. Defenders~typically set a lower limit of the BA drop, $\delta$, to assess~whether~the~defended model is acceptable. Prior works generally consider~an~effective defense to have a BA drop of no more than~10\% \cite{wu2022backdoorbench,liu2018fine,zeng2022adversarial}, and thereby we set $\delta$ to -10\%. After eliminating the backdoor via a defense, if BA still remains high (i.e., the BA drop is within~the acceptable limit $\delta$), defenders will believe that the model has~been successfully defended and accept the defended model. In this~case, the smaller the drop in ASR compared to the undefended model, the stronger the survivability of the attack. Conversely, if BA~becomes low (i.e., the BA drop exceeds $\delta$), defenders will tend~to~not use the ineffective defense and still adopt the undefended model. In this case, the larger the drop in BA, the more likely to adopt~the undefended model, and the stronger survivability of the attack; and meanwhile, the smaller the drop in ASR, the stronger~the~survivability of the attack. Overall, we expect the defended model~either meets the effectiveness goal or fails to meet the utility goal. Based on this intuition, we define ASuR by Eq.~\ref{eq:asur}.
\begin{equation}\label{eq:asur}
\small
    \begin{aligned}
    \text{ASuR} &= \begin{cases}
        \alpha_1 \mathop{Scaler}(\Delta \text{ASR}) + \alpha_2\mathop{Scaler}(\Delta \text{BA}) \ \ \ , \text{if}\ \Delta \text{BA} \ge \delta \\
        \alpha_1 \mathop{Scaler}(\Delta \text{ASR}) + \alpha_2\mathop{Scaler}(- \Delta \text{BA}) \ ,\text{if}\ \Delta  \text{BA} < \delta
       \end{cases}\\
    & = \begin{cases}
        0.95  \text{ASR}_a + 0.05  \frac{\Delta \text{BA}-\delta}{1-\text{BA}_b-\delta}\ ,\text{if}\ \Delta \text{BA} \ge \delta \\
        0.50 \text{ASR}_a + 0.50 \frac{\delta - \Delta \text{BA}}{\text{BA}_b+\delta} \ \ \ \ ,\text{if}\ \Delta  \text{BA} < \delta
       \end{cases}
    \end{aligned}
\end{equation}
where $\Delta \text{ASR}$ and $\Delta \text{BA}$ respectively denote the drop in ASR~and~BA before and after a defense is applied. $\text{ASR}_a$ and $\text{BA}_b$ respectively denote the ASR after a defense is applied and the BA~before a defense is applied. We employ Min-Max Scaler as the~$\mathop{Scaler}(\cdot)$ function. Specifically, when $\Delta{BA}$ is above $\delta$, the drop in BA~is~usually considered as insignificant. Therefore, we set $\alpha_1$ to 0.95~and~$\alpha_2$ to 0.05 to emphasize the drop in ASR. Here, $\Delta \text{BA}-\delta$ and $1-\text{BA}_b-\delta$ represent the increase and maximum increase in BA, respectively. When $\Delta{BA}$ is below $\delta$, we set $\alpha_1$ to 0.50 and $\alpha_2$ to 0.50~to~comprehensively trade-off the drop in ASR and BA. Here, $\delta - \Delta \text{BA}$~and $\text{BA}_b+\delta$ represent the decrease and maximum decrease in BA,~respectively. Therefore, a higher ASuR indicates a stronger survivability of an attack against a defense.
\end{itemize}

\textbf{Environment.} All our experiments were conducted on a Linux server with two 16-cores 32-threads Intel(R) Xeon(R) Silver 4314 CPUs as well as two NVIDIA GeForce RTX 3090 GPUs, running~the Ubuntu 18.04.6 LTS x86\_64 operating system.

\subsection{Capability Preservation Evaluation (RQ1)}\label{sec:rq1}

\begin{table}[!t]
    \caption{Evaluation results of utility and effectiveness.} 
    \label{tbl:utility and effectiveness}
    \vspace{-10pt}
    \scriptsize
    \centering
    \begin{tabular}{c p{0.67cm}p{0.32cm} p{0.73cm} p{0.32cm} p{0.73cm} p{0.32cm} p{0.73cm} p{0.32cm}}
        \toprule
        \multirow{2}{*}{Attack} & \multicolumn{2}{c}{Original} & \multicolumn{2}{c}{\tool-Enhanced} & \multicolumn{2}{c}{Original} & \multicolumn{2}{c}{\tool-Enhanced} \\ 
        \cmidrule(lr){2-3} \cmidrule(lr){4-5} \cmidrule(lr){6-7} \cmidrule(lr){8-9}
        & BA/BAD & ASR & BA/BAD & ASR & BA/BAD & ASR & BA/BAD & ASR\\ 
        \midrule 
        & \multicolumn{4}{c}{VGG19-BN + CIFAR-10} & \multicolumn{4}{c}{VGG19-BN + CIFAR-100} \\\cmidrule(lr){2-5}\cmidrule(lr){6-9}
        BadNets & 90.78/1.45 & 95.06 & 90.12/2.11 & 96.03 & 60.93/4.54 & 89.06 & 58.29/7.18 & 90.24  \\
        Blend & 91.99/0.24 & 99.71 & 91.55/0.68 & 99.71 & 64.88/0.59 & 98.97 & 64.82/0.65 & 98.96\\
        TrojanNN & 91.44/0.79 & 100.00 & 91.85/0.38 & 99.99 & 65.00/0.47 & 99.99 & 65.77/${-}$0.30 & 100.00 \\
        LC & 83.48/8.75 & 99.46 & 83.52/8.71 & 99.82 & 65.58/${-}$0.11 & 17.15 & 66.49/${-}$1.02 & 55.76\\
        SSBA & 91.47/0.76 & 95.61 & 90.59/1.64 & 95.34 & 64.09/1.38 & 94.99 & 62.33/3.14 & 95.39\\ 
        Inputaware & 89.51/2.72 & 90.31 & 88.67/3.56 & 95.52 & 58.24/7.23 & 89.54 & 58.00/7.47 & 86.56 \\
        WaNet & 85.17/7.06 & 99.09 & 85.45/6.78 & 98.47 & 55.13/10.34 & 98.45 & 50.73/14.74 & 98.05 \\
        Adap-Blend & 92.10/0.13 & 71.90 & 92.07/0.16 & 70.12 & 68.49/${-}$3.02 & 51.93 & 68.54/${-}$3.07 & 71.44 \\\midrule
        Avg & \textbf{89.49/2.74} & 93.89 & 89.23/3.00 & \textbf{94.38} & \textbf{62.79/2.68} & 80.01 & 61.87/3.60 & \textbf{87.05} \\
        \midrule 
        & \multicolumn{4}{c}{VGG19-BN + GTSRB} & \multicolumn{4}{c}{PreActResNet18 + CIFAR-10} \\\cmidrule(lr){2-5}\cmidrule(lr){6-9}
        BadNets & 97.49/0.63 & 94.81 & 97.97/0.15 & 95.34 & 91.41/2.42 & 94.78 & 89.94/3.89 & 95.90 \\
        Blend & 97.85/0.27 & 99.94 & 98.36/${-}$0.24 & 99.98 & 93.69/0.14 & 99.87 & 93.17/0.66 & 99.86 \\
        TrojanNN & 98.09/0.03 & 100.00 & 98.35/${-}$0.23 & 100.00 & 93.67/0.16 & 100.00 & 93.76/0.07 & 99.99 \\
        LC & 97.82/0.30 & 63.03 & 97.65/0.47 & 63.74 & 84.95/8.88 & 93.36 & 85.18/8.65 & 92.14 \\
        SSBA & 97.75/0.37 & 99.55 & 97.41/0.71 & 99.24 & 93.07/0.76 & 97.53 & 93.33/0.50 & 98.08\\
        Inputaware & 94.96/3.16 & 77.17 & 96.37/1.75 & 90.86 & 91.18/2.65 & 98.70 & 91.40/2.43 & 95.00 \\
        WaNet & 94.78/3.34 & 98.12 & 94.84/3.28 & 93.83 & 90.12/3.71 & 85.99 & 90.05/3.78 & 96.91\\
        Adap-Blend & 97.21/0.91 & 70.54 & 96.48/1.64 & 71.11 & 93.68/0.15 & 74.63 & 93.10/0.73 & 74.28 \\\midrule
        Avg & 96.99/1.13 & 87.90 & \textbf{97.18/0.94} & \textbf{89.26} & \textbf{91.47/2.36} & 93.11 & 91.24/2.59 & \textbf{94.02} \\
        \bottomrule 
    \end{tabular}
\end{table}

\begin{table}[!t]
    \caption{Evaluation results of trigger stealthiness.}
    \label{tbl:stealthiness}
    \vspace{-10pt}
    \footnotesize
    \begin{tabular}{cp{0.82cm}p{0.82cm}p{0.82cm}p{0.82cm}p{0.82cm}p{0.82cm}}
        \toprule
        \multirow{2}{*}{Attack} & \multicolumn{3}{c}{Original} & \multicolumn{3}{c}{\tool-Enhanced} \\  
        \cmidrule(lr){2-4}\cmidrule(lr){5-7} 
        & PSNR$\uparrow$ & $l^\infty \downarrow$ & LPIPS$\downarrow$ & PSNR$\uparrow$ & $l^\infty \downarrow$ & LPIPS$\downarrow$ \\
        \midrule 
        BadNets & 27.308 & 169.792 & 0.0010 & 27.308 & 169.792 & 0.0010 \\
        Blend & 20.655 & 47.969 & 0.0353 & 20.655 & 47.969 & 0.0353\\
        TrojanNN & 19.385 & 217.477 & 0.0431 & 19.385 & 217.477 & 0.0431 \\
        LC & 24.465 & 204.027 & 0.0090 & 24.465 & 204.027 & 0.0090\\
        SSBA & 24.847 & 82.982 & 0.0139 & 24.847 & 82.982 & 0.0139\\
        Inputaware & 21.934 & 203.454 & 0.0687 & 21.906 & 200.111 & 0.0430\\
        WaNet & 31.164 & 45.164 & 0.0055 & 31.164 & 45.164 & 0.0055\\
        Adap-Blend & 21.019 & 48.844 & 0.0488 & 21.019 & 48.844 & 0.0488\\ 
        \bottomrule
    \end{tabular}
\end{table}

We use \tool to enhance eight attacks, and evaluate how \tool~affects the capability of original attacks in terms of two aspects. First, we measure \tool's impact on the utility and effectiveness~by~comparing the BA and ASR of the original attacks and \tool-enhanced attacks. Second, we explore \tool's impact on the specific~attack characteristics, i.e., trigger stealthiness and latent inseparability.

\textbf{Utility and Effectiveness.} Table~\ref{tbl:utility and effectiveness} reports the BA, BAD~and~ASR of the original attacks and \tool-enhanced attacks on the four~classification tasks. On average, original attacks achieve a BAD of \todo{2.23\%}, whereas \tool-enhanced attacks have a BAD of \todo{2.53\%}. Meanwhile, original attacks have an ASR of \todo{88.73\%}, while \tool-enhanced~attacks achieve an ASR of \todo{91.18\%}. Therefore, \tool helps to slightly improve ASR by \todo{2.45\%} at a cost of slight drop in BA by \todo{0.30\%}.

\textbf{Trigger Stealthiness.} We use three widely used metrics,~i.e.,~peak signal-to-noise ratio (PSNR)~\cite{huynh2008scope}, $l^\infty$~\cite{larsen2005introduction}, and learned~perceptual image patch similarity (LPIPS)~\cite{zhang2018unreasonable}, to measure the stealthiness of triggers. Table~\ref{tbl:stealthiness} shows the results of the original~attacks~and~\tool-enhanced attacks on VGG19-BN using CIFAR-10; and the results~on~the other three tasks are similar and thus are omitted. \tool-enhanced attacks  have almost the same trigger stealthiness as original attacks.

\begin{figure}[!t]
    \centering
    \subcaptionbox{\footnotesize Blend}{\includegraphics[width=.33\linewidth]{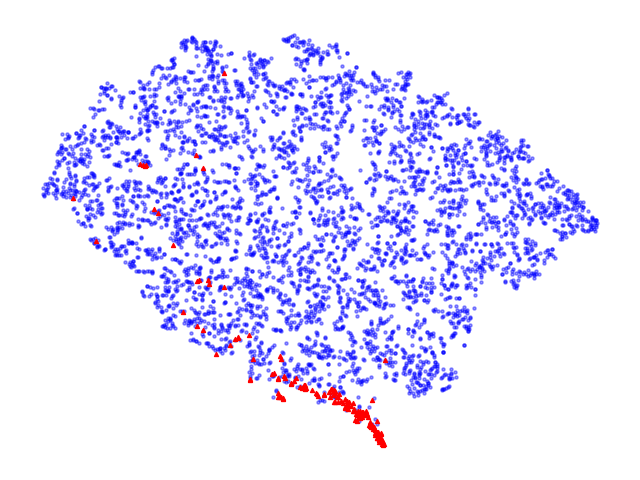}\vspace{-5pt}}\hfill
    \subcaptionbox{\footnotesize Adap-Blend}{\includegraphics[width=.33\linewidth]{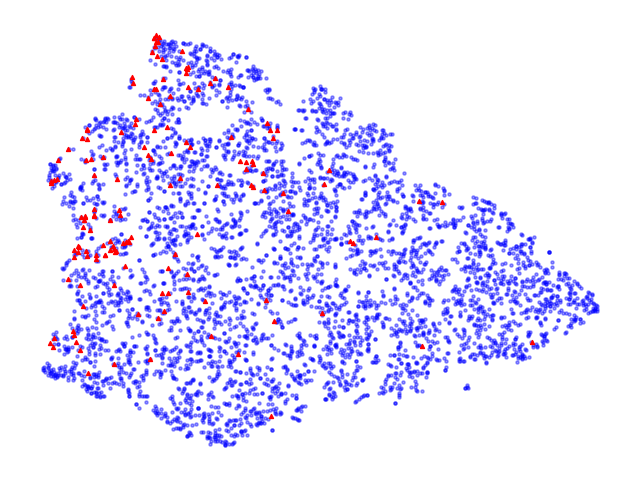}\vspace{-5pt}}\hfill
    \subcaptionbox{\footnotesize \tool-Adap-Blend}{\includegraphics[width=.33\linewidth]{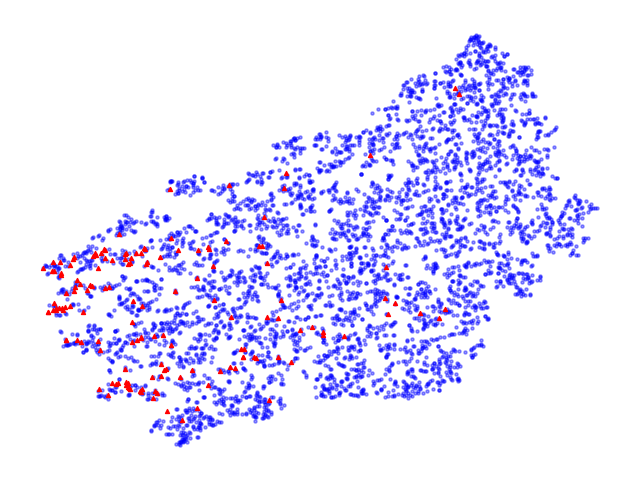}\vspace{-5pt}}
    \vspace{-8pt}
    \caption{T-SNE visualization of latent inseparability. Benign and poisoned samples are respectively blue and red points.}
    \label{fig:t-sne}
\end{figure}

\textbf{Latent Inseparability.} Adap-Blend has a specific characteristic of latent inseparability that is missing from other attacks. Therefore, we analyze whether \tool affects this characteristic. Figure~\ref{fig:t-sne} plots the latent representations of poisoned and benign samples for Blend, Adap-Blend and \tool-enhanced Adap-Blend (i.e., \tool-Adap-Blend) on VGG19-BN using CIFAR-10, visualized by T-SNE~\cite{van2008visualizing}; and the results on the other three tasks are similar and hence are omitted. Notable latent separation between benign and poisoned samples~is observed for Blend, but the poisoned and benign~samples~mix~with each other for both Adap-Blend and \tool-Adap-Blend. 

Moreover, we explore whether \tool affects Adap-Blend's resistance to four data distribution-based defenses, i.e.,  Spectral~\cite{tran2018spectral}, AC~\cite{chen2018detecting}, SCAn~\cite{tang2021demon} and SPECTRE~\cite{hayase2021spectre}. Table~\ref{tbl:detection} presents the results~on VGG19-BN using CIFAR-10; and the results on the other~three tasks are similar and thereby are omitted. The columns \textit{Eliminate}~and~\textit{Sacrifice} respectively denote the ratio of poisoned samples~that~are~eliminated and the ratio of benign samples that are mistakenly removed (i.e., sacrificed). Spectral and SPECTRE eliminate most of the poisoned samples with a negligible sacrifice of benign samples when defending Blend. Differently, both Adap-Blend and \tool-Adap-Blend evade all the four defenses with a low elimination rate~of~poisoned samples and a low sacrifice rate of benign samples.

\begin{table}[!t]
    \caption{Data distribution-based defenses against attacks.}
    \label{tbl:detection}
    \footnotesize
    \vspace{-10pt}
    \centering
    \begin{tabular}{cp{0.81cm}p{0.81cm}p{0.81cm}p{0.81cm}p{0.81cm}p{0.81cm}}
        \toprule
        \multirow{2}{*}{Defense} & \multicolumn{2}{c}{Blend} & \multicolumn{2}{c}{Adap-Blend} & \multicolumn{2}{c}{\tool-Adap-Blend} \\  
        \cmidrule(lr){2-3}\cmidrule(lr){4-5} \cmidrule(lr){6-7}
        & Eliminate & Sacrifice & Eliminate & Sacrifice & Eliminate & Sacrifice\\
        \midrule 
        Spectral & 62.86 & 4.42 & 0.67 & 4.51 & 14.67 & 4.47\\
        AC & 0.00 & 0.00 & 1.33 & 3.17 & 0.00 & 0.00 \\
        SCAn & 0.00 & 0.00 & 0.00 & 4.37 & 0.00 & 4.39\\
        SPECTRE& 86.47 & 0.23 & 0.00 & 0.45 &0.00 & 0.45\\
        \bottomrule
    \end{tabular}
\end{table}

\textbf{Summary.}  \tool preserves the capability of original attacks.~In detail, \tool keeps the utility, effectiveness and trigger stealthiness of original attacks, and also keeps the latent inseparability and the resistance to data distribution-based defenses of Adap-Blend.

\subsection{Survivability Evaluation (RQ2)}\label{RQ2}

\begin{table*}[!t]
    \caption{Evaluation results of survivability on VGG19-BN using CIFAR-10 (O./V. denotes the original/\tool-enhanced attack).}
    \label{tbl:survivability}
    \centering
    \vspace{-10pt}
    \footnotesize
    \begin{tabular}{cc *{18}{p{0.51cm}}}
        \toprule
        \multirow{2}{*}{Defense} & \multirow{2}{*}{Metric} & 
            \multicolumn{2}{c}{BadNets} &
            \multicolumn{2}{c}{Blend} & 
            \multicolumn{2}{c}{TrojanNN} &
            \multicolumn{2}{c}{LC} & 
            \multicolumn{2}{c}{SSBA} & 
            \multicolumn{2}{c}{Inputaware} & 
            \multicolumn{2}{c}{WaNet} & 
            \multicolumn{2}{c}{Adap-Blend} & 
            \multicolumn{2}{c}{{Avg}} \\
         \cmidrule(lr){3-4}\cmidrule(lr){5-6}\cmidrule(lr){7-8}\cmidrule(lr){9-10}\cmidrule(lr){11-12}\cmidrule(lr){13-14}\cmidrule(lr){15-16}\cmidrule(lr){17-18}\cmidrule(lr){19-20}
         & & O. & V. & O. & V. &  O. & V. &  O. & V. &  O. & V. &  O. & V. &  O. & V. &  O. & V. & {O.} & {V.} \\ 
        \midrule 
        \multirow{3}{*}{FT} & BA & 88.80 & 90.30 & 89.82 & 90.81 & 89.61 & 89.95 & 88.51 & 90.37 & 89.50 & 90.38 & 91.33 & 90.63 & 91.36 & 91.27 & 90.87 & 91.99 & 89.98 & \textbf{90.71} \\
        & ASR & 5.93 & 71.77 & 86.08 & 96.18 & 7.94 & 96.97 & 26.59 & 85.72 & 62.88 & 86.53 & 2.42 & 70.56 & 1.97 & 42.46 & 60.32 & 67.52 & 31.77 & \textbf{77.21} \\
        & ASuR & 7.72 & 70.74 & 83.95 & 93.88 & 9.74 & 94.35 & 28.09 & 84.62 & 61.90 & 84.73 & 5.18 & 69.84 & 5.13 & 43.56 & 59.75 & 66.91 & 32.68 & \textbf{76.08} \\
        \midrule 
        \multirow{3}{*}{I-BAU} & BA & 85.34 & 89.41 & 86.45 & 86.48 & 85.84 & 88.61 & 86.55 & 86.89 & 87.66 & 88.80 & 88.93 & 88.24 & 89.12 & 89.58 & 90.48 & 89.87 & 87.55 & \textbf{88.48} \\
        & ASR & 0.94 & 66.73 & 30.77 & 94.71 & 10.93 & 31.47 & 47.71 & 73.71 & 7.34 & 8.20 & 80.98 & 88.20 & 28.29 & 27.33 & 20.81 & 50.66 & 28.47 & \textbf{55.13} \\
        & ASuR & 2.08 & 65.73 & 30.47 & 91.31 & 11.57 & 31.76 & 47.79 & 72.55 & 8.64 & 9.90 & 79.23 & 86.03 & 29.68 & 28.84 & 22.11 & 50.30 & 28.95 & \textbf{54.55} \\
        \midrule 
        \multirow{3}{*}{NAD} & BA & 88.15 & 89.62 & 88.16 & 90.54 & 88.45 & 88.88 & 87.20 & 89.91 & 88.93 & 90.22 & 90.99 & 90.35 & 91.01 & 91.23 & 90.73 & 91.45 & 89.20 & \textbf{90.28} \\
        & ASR & 27.86 & 79.00 & 79.82 & 98.82 & 7.72 & 95.38 & 62.17 & 82.79 & 58.07 & 82.97 & 2.21 & 52.42 & 0.84 & 41.22 & 66.49 & 66.31 & 38.15 & \textbf{74.86} \\
        & ASuR & 28.38 & 77.44 & 77.54 & 96.32 & 9.22 & 92.55 & 61.65 & 81.75 & 57.18 & 81.30 & 4.90 & 52.54 & 3.99 & 42.37 & 65.58 & 65.61 & 38.55 & \textbf{73.73} \\
        \midrule 
        \multirow{3}{*}{NC} & BA & 88.49 & 89.92 & 91.99 & 91.55 & 89.34 & 91.85 & 88.24 & 90.45 & 91.47 & 90.59 & 90.52 & 88.38 & 90.42 & 85.45 & 90.59 & 92.07 & \textbf{90.13} & 90.03 \\
        & ASR & 14.19 & 67.73 & 99.71 & 99.71 & 6.32 & 99.99 & 30.43 & 91.79 & 95.61 & 95.34 & 2.91 & 93.47 & 0.39 & 98.47 & 51.84 & 70.12 & 37.67 & \textbf{89.58} \\
        & ASuR & 15.49 & 66.81 & 97.50 & 97.43 & 8.13 & 97.75 & 31.69 & 90.40 & 93.53 & 93.15 & 5.45 & 91.07 & 3.44 & 95.58 & 51.62 & 69.40 & 38.36 & \textbf{87.70} \\
        \midrule 
        \multirow{3}{*}{FP} & BA & 89.06 & 90.14 & 90.17 & 90.51 & 89.79 & 89.86 & 89.26 & 90.64 & 89.12 & 90.05 & 91.43 & 90.49 & 90.99 & 91.42 & 91.36 & 91.81 & 90.15 & \textbf{90.61} \\
        & ASR & 32.89 & 76.18 & 91.44 & 99.67 & 7.74 & 95.77 & 70.12 & 93.56 & 60.58 & 87.36 & 31.19 & 51.41 & 1.30 & 61.23 & 49.19 & 65.12 & 43.06 & \textbf{78.79} \\
        & ASuR & 33.40 & 74.89 & 89.14 & 97.11 & 9.60 & 93.19 & 69.59 & 92.11 & 59.62 & 85.43 & 32.54 & 51.61 & 4.42 & 61.42 & 49.32 & 64.58 & 43.45 & \textbf{77.54} \\
        \midrule 
        \multirow{3}{*}{BNP} & BA & 90.10 & 89.66 & 91.17 & 90.73 & 90.38 & 91.23 & 83.05 & 83.08 & 90.98 & 90.24 & 90.00 & 84.81 & 66.79 & 25.94 & 91.22 & 91.73 & \textbf{86.71} & 80.93 \\
        & ASR & 95.14 & 95.80 & 99.19 & 99.37 & 100.00 & 100.00 & 99.78 & 99.89 & 95.63 & 95.20 & 3.44 & 3.64 & 55.22 & 93.92 & 71.43 & 71.87 & 77.48 & \textbf{82.46} \\
        & ASuR & 92.81 & 93.41 & 96.78 & 96.89 & 97.41 & 97.58 & 96.60 & 96.70 & 93.41 & 92.93 & 5.83 & 4.90 & 33.18 & 79.77 & 70.41 & 70.97 & 73.30 & \textbf{79.14} \\
        \midrule 
        \multirow{3}{*}{CLP} & BA & 89.11 & 73.11 & 87.09 & 89.78 & 89.01 & 90.95 & 81.31 & 82.28 & 89.49 & 82.97 & 89.86 & 84.81 & 71.37 & 86.43 & 91.63 & 88.02 & \textbf{86.11} & 84.79 \\
        & ASR & 34.03 & 3.02 & 79.58 & 97.81 & 74.27 & 92.04 & 0.01 & 64.42 & 88.84 & 95.89 & 43.91 & 3.92 & 81.43 & 42.94 & 39.68 & 40.87 & \textbf{55.22} & 55.11 \\
        & ASuR & 34.50 & 5.88 & 77.02 & 95.15 & 72.60 & 89.94 & 1.49 & 62.85 & 86.56 & 91.71 & 44.24 & 5.16 & 43.24 & 43.03 & 40.36 & 40.49 & 50.00 & \textbf{54.28} \\
        \midrule 
        \multirow{3}{*}{NPD} & BA & 90.11 & 89.92 & 90.92 & 90.84 & 91.40 & 91.16 & 86.46 & 87.23 & 89.96 & 90.00 & 88.49 & 87.77 & 89.73 & 88.71 & 90.99 & 91.44 & \textbf{89.76} & 89.63 \\
        & ASR & 1.30 & 65.48 & 6.17 & 91.71 & 69.08 & 99.99 & 8.39 & 21.89 & 3.57 & 71.41 & 50.12 & 15.30 & 6.13 & 12.80 & 33.20 & 62.26 & 22.24 & \textbf{55.11} \\
        & ASuR & 3.66 & 64.67 & 8.34 & 89.64 & 68.31 & 97.56 & 10.42 & 23.38 & 5.68 & 70.26 & 49.81 & 16.67 & 8.76 & 14.86 & 34.02 & 61.76 & 23.62 & \textbf{54.85} \\
        \midrule 
        \multirow{3}{*}{Avg} & BA & \textbf{88.64} & 87.76 & 89.47 & \textbf{90.16} & 89.23 & \textbf{90.31} & 86.32 & \textbf{87.61} & \textbf{89.64} & 89.16 & \textbf{90.19} & 88.19 & \textbf{85.10} & 81.25 & 90.98 & \textbf{91.05} & \textbf{88.70} & 88.18 \\
        & ASR & 26.54 & \textbf{65.71} & 71.59 & \textbf{97.25} & 35.50 & \textbf{88.95} & 43.15 & \textbf{76.72} & 59.06 & \textbf{77.86} & 27.15 & \textbf{47.36} & 21.95 & \textbf{52.55} & 49.12 & \textbf{61.84} & 41.76 & \textbf{71.03} \\
        & ASuR & 27.25 & \textbf{64.95} & 70.09 & \textbf{94.72} & 35.82 & \textbf{86.83} & 43.41 & \textbf{75.55} & 58.31 & \textbf{76.18} & 28.40 & \textbf{47.23} & 16.48 & \textbf{51.18} & 49.15 & \textbf{61.25} & 41.12 & \textbf{69.73} \\                       
        \bottomrule
    \end{tabular}
\end{table*}

We use eight defenses to defend the eight original attacks~and~eight \tool-enhanced attacks, and investigate how \tool enhances~the attack survivability by comparing the BA, ASR and ASuR~of~the~original and \tool-enhanced attacks after the defenses are applied.~As different defense follow different strategies, we conduct~detailed~analysis for different types of defenses respectively.

\textbf{Results on VGG19-BN Using CIFAR-10.} Table~\ref{tbl:survivability} reports the attack survivbility comparison on VGG19-BN using CIFAR-10. After the defenses are applied, the original attacks still maintain a high~BA (i.e., \todo{88.70\%} on average), while their ASR~drops~to \todo{41.76\%}~averagely. Notably, \todo{32.8\%} of the original attacks have an ASR that is lower than 10\% after certain defenses are applied. On the contrary,~the~\tool-enhanced attacks achieve an average BA of \todo{88.18\%} and an average ASR of \todo{71.03\%}. Notably, \todo{39.1\%} of the \tool-enhanced attacks~have an ASR that is higher than 90\% after certain defenses~are~applied. In that sense, \tool enhances ASR by \todo{29.27\%} with a slight decrease~of BA by \todo{0.52\%}. With respect to the comprehensive metric~of~ASuR, \tool significantly enhances the attack survivability against~various defenses, i.e., an average increase in ASuR~from~\todo{41.12\%}~to~\todo{69.73\%}.

\textbf{Attack Survivability Against FT/I-BAU/NAD.}
As shown in Table~\ref{tbl:survivability}, \tool significantly reduces the effectiveness of these three model unlearning-based defenses, therefore enhancing~the~survivability of attacks in almost all configurations. Numerically, we observe an average increase in ASuR against FT by \todo{43.40\%}, I-BAU by \todo{25.60\%} and NAD by \todo{35.18\%}. This is due to our tight coupling~between the backdoor and the model's classification performance.

\begin{figure}[!t]
    \centering
    \includegraphics[width=.45\textwidth]{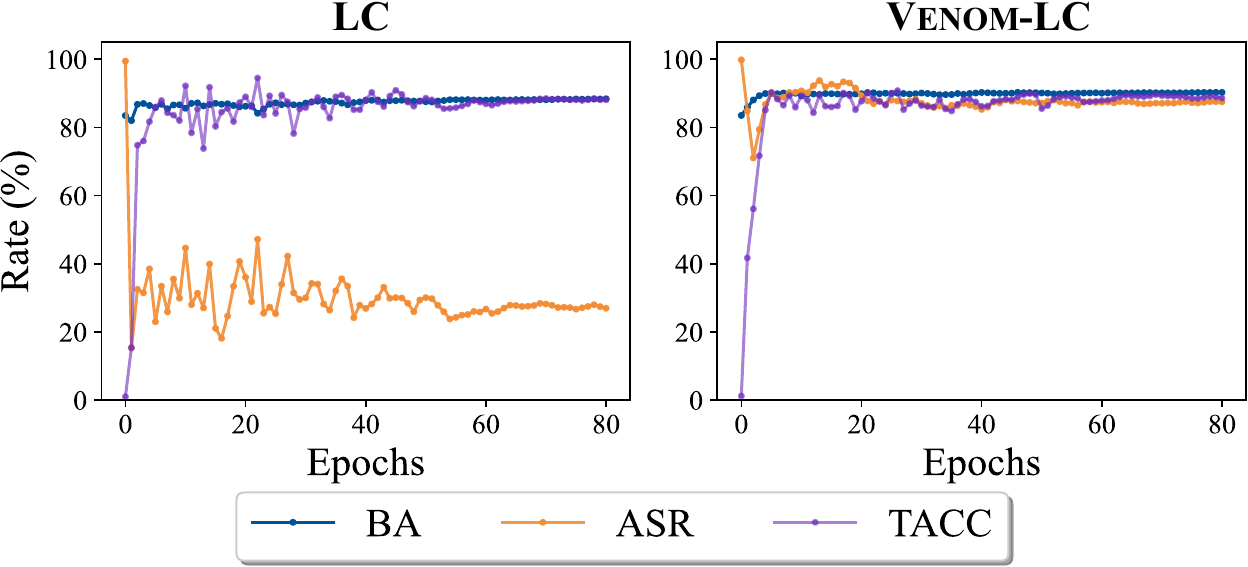}
    \vspace{-5pt}
    \caption{Defense process of FT against LC and \tool-LC.}
    \label{fig:LC_FT}
\end{figure}

Moreover, we take the LC attack against the FT defense~as~an~example for a deeper insight. LC is a clean-label attack, and it causes the backdoored model to completely disregard the intrinsic features of target-class samples and instead treat the trigger as the target class, thus sacrificing target-class accuracy (TACC) to ensure the success of the attack. Therefore, as shown by the defense process of FT in Figure~\ref{fig:LC_FT}, LC and \tool-enhanced LC (i.e., \tool-LC) have a very low TACC of \todo{1.00\%} and \todo{1.20\%} respectively, before~FT~is~applied. FT aims to force the defended model's attention back to the samples themselves, effectively rendering the backdoored model ineffective. In practice, as shown by the defense process of FT against LC, there is a decrease of ASR from \todo{99.46\%} to \todo{26.59\%} and an increase of BA from \todo{83.48\%} to \todo{88.51\%}, and TACC restores to a high~value of \todo{88.20\%}. However, as shown by the defense process of FT against \tool-LC, as TACC increases, ASR exhibits only minor fluctuations before stabilizing at a high value of \todo{85.72\%}. This result indicates that, due to the tight coupling between the backdoor and the model's classification performance, repairing the model's classification performance does not lead to the forgetting of the backdoor, instead, it actually helps to maintain the success of the attack.

\textbf{Attack Survivability Against NC.} 
As shown in Table~\ref{tbl:survivability}, \tool-enhanced attacks consistently maintain a high ASR after NC~is~applied. Specifically, only \tool-BadNets and \tool-Adap-Blend have an ASR around 70\%, and the remaining \tool-enhanced~attacks achieve an ASR over 90\%, which indicates the failure of NC. Compared to the original attacks, \tool-enhanced attacks exhibit an average increase of \todo{51.91\%} in ASR and \todo{49.34\%} in ASuR.

\begin{figure}[!t]
    \centering
    \subcaptionbox{\footnotesize Original Trigger in BadNets}{\setlength{\fboxrule}{0.01pt} \setlength{\fboxsep}{0pt} \fbox{\includegraphics[width=.295\linewidth]{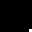}}\vspace{-5pt}}\hfill
    \subcaptionbox{\footnotesize Synthesized Trigger for BadNets}{\setlength{\fboxrule}{0.01pt} \setlength{\fboxsep}{0pt} \fbox{\includegraphics[width=.295\linewidth]{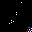}}\vspace{-5pt}}\hfill
    \subcaptionbox{\footnotesize Synthesized Trigger for \tool-BadNets}{\setlength{\fboxrule}{0.01pt} \setlength{\fboxsep}{0pt} \fbox{\includegraphics[width=.295\linewidth]{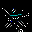}}\vspace{-5pt}}
    \vspace{-8pt}
    \caption{The original trigger in BadNets as well~as~NC's~synthesized triggers for BadNets and \tool-BadNets.}
    \label{fig:NC-triggers}
\end{figure}

Besides, we take the BadNets attack against the NC defense~as~an example to show how \tool evades NC. NC detects the poisoned target label, and synthesizes the trigger before eliminating~the~backdoor via model~unlearning, and it requires that the synthesized trigger should be roughly similar to the original~trigger~(at~least~for~BadNets) in terms~of~shape and location~\cite{wang2019neural}. As shown in Figure~\ref{fig:NC-triggers},~the synthesized trigger for BadNets appears to be similar~to~the original trigger, as both focus on the bottom-right corner of the image.~However, the synthesized trigger for \tool-BadNets differs~significantly from the original trigger in terms of shape and location.

In fact, except for \tool-BadNets and \tool-LC, all \tool-enhanced attacks evade the label detection in NC; i.e., NC considers the backdoored model as a clean model. Although NC correctly~detects the poisoned label in \tool-BadNets and \tool-LC, it still fails to eliminate the backdoor because of the different shapes and locations between the synthesized and original triggers.

\textbf{Attack Survivability Against FP/BNP/CLP/NPD.}
As shown in Table~\ref{tbl:survivability}, \tool significantly enhance ASuR against FP and NPD, achieving an increase of \todo{34.09\%} and \todo{31.23\%} respectively. While in the case of BNP and CLP, the enhancement is comparatively limited, with an ASuR increase of \todo{5.84\%} and \todo{4.28\%}, respectively.~As~most~of the original attacks already exhibit high resistance against BNP, the enhancement of \tool is therefore limited in this particular case. Besides, \tool fails to enhance the survivability of BadNets~and~Inputaware against CLP, leading to a limited ASuR improvement.

\begin{figure}[!t]
    \centering
    \includegraphics[width=.41\textwidth]{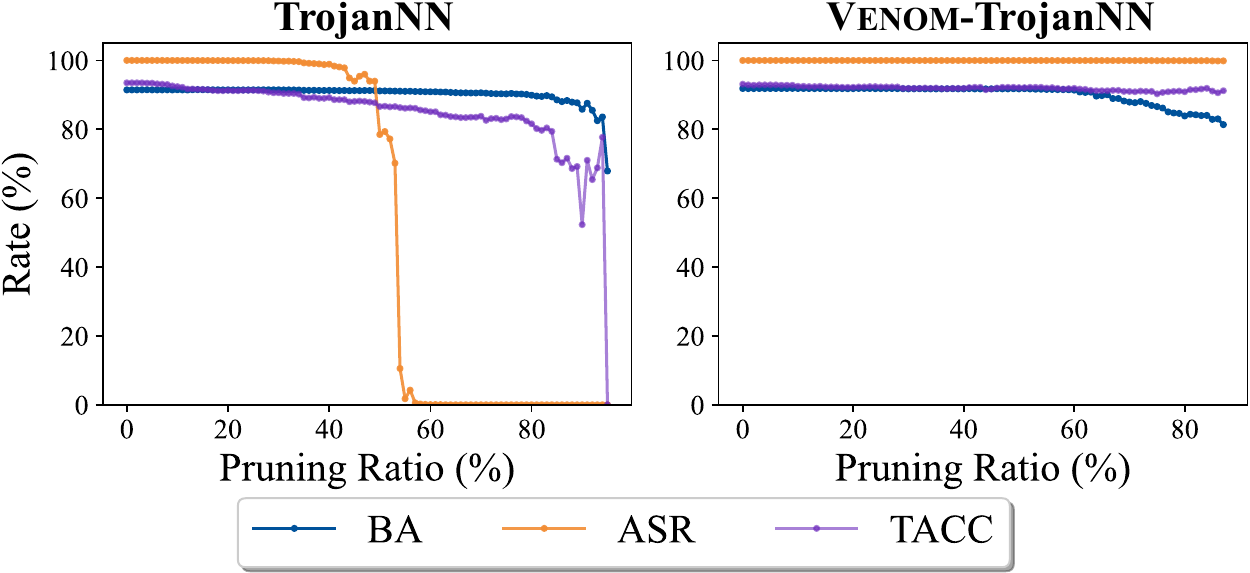}
    \vspace{-5pt}
    \caption{Pruning process of FP against TrojanNN and \tool-TrojanNN with respect to the pruning ratio.}
    \label{fig:Pruning-based}
\end{figure}

We take~the TrojanNN attack against the FP defense~as~an~example for~a~deeper~investigation. FP eliminates the backdoor~by~pruning backdoor-related neurons of the selected layer. As illustrated by the pruning process of FP against TrojanNN and \tool-TrojanNN~in Figure~\ref{fig:Pruning-based},~ASR~of~\tool-TrojanNN remains around 100\% even~with a pruning ratio of 90\%, but ASR of the original TrojanNN decreases significantly, with the backdoor completely eliminated~when~the~pruning ratio reaches 60\%.
\tool-enhanced attacks evade FP because \tool embeds~the~backdoor into the neurons which are crucial for the classification of benign samples, rendering the decision path of poisoned samples overlapping with that of benign samples.

\textbf{Results on the Other Three Tasks.} For VGG19-BN on CIFAR-100, after the defenses are applied, \tool helps existing attacks to averagely enhance ASR by \todo{30.62\%} with an average decrease~in~BA~of \todo{0.24\%}. Overall, \tool enhances the average attack survivability in terms of ASuR from \todo{32.91\%} to \todo{61.85\%}. Consistently, for VGG19-BN on GTSRB, \tool causes an average increase in attack survivability from \todo{44.92\%} to \todo{63.97\%}. For PreActResNet18 on CIFAR-10,~\tool~increases the average attack survivability from \todo{37.44\%} to \todo{54.24\%}.
Table~\ref{tbl:vgg19 on cifar100}, \ref{tbl:vgg19 on gtsrb} and \ref{tbl:resnet on cifar10} in Appendix~\ref{Appendix:Results of Other Datasets and Models} report the detailed attack survivability comparison across attacks and defenses.

\textbf{Summary.} \tool significantly enhances the original attacks' survivability against model reconstruction-based defenses, leading to an average increase in ASuR from \todo{39.10\%} to \todo{62.45\%}.

\subsection{Ablation Study (RQ3)}
\label{sec:Influencing factors}

\begin{figure*}[!t]
    \centering
    \includegraphics[width=.89\textwidth]{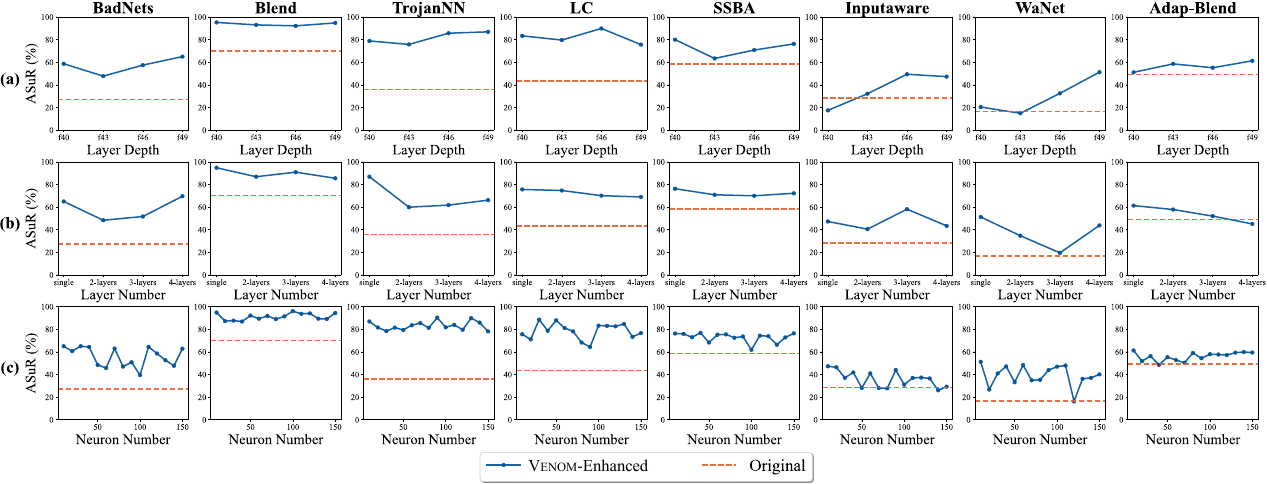}
    \vspace{-7pt}
    \caption{Impact of (a) layer depth, (b) layer number and (c) neuron number on the survivability of \tool-enhanced attacks.}
    \label{fig:impacts}
\end{figure*}

We evaluate the impact of several factors in the key TCDP generation step in \tool, including the depth of the selected layer,~the~number of selected layers, and the number of selected neurons, on~the~survivability of \tool-enhanced attacks. Further, we ablate each~step~in \tool, i.e., micro-training, TCDP generation, and optimized binary-task training to measure the contribution~of each step. We conduct all the experiments on VGG19-BN using CIFAR-10. 
For each attack, we report the average ASuR against eight defenses.

\textbf{Impact of Layer Depth.} 
We perform TCDP generation~on~each convolutional layer. However, the shallow layers have an empty TCDP, meaning that they are not directly crucial~to target-class~classification. Thus, we select the last four convolutional layers in the deep layers with the name of $features.40$, $features.43$,~$features.46$ and $features.49$ (abbreviated as f40, f43, f46 and f49), which contain at least 10 neurons in the generated TCDP. Figure~\ref{fig:impacts}(a) reports the impact of different layer depths on the average ASuR. Here we set the average ASuR of the original attacks as baselines. Overall,~except for f40 for \tool-Inputaware and f43 for \tool-WaNet,~the~four layers all exhibit better ASuR than baselines, and as the layer depth increases, ASuR mostly has an increasing trend. Therefore, we use the last convolutional layer (i.e., f49) as the target single layer.

\textbf{Impact of Layer Number.}
In TCDP generation, we select crucial neurons from one single layer. However, crucial neurons may~be distributed across different layers. Hence, we conduct experiments on the single-layer setting and multi-layer settings. For the single-layer setting, we select f49 as the target single layer. For multi-layer settings, we use the above four deep layers to form paths~of~different lengths, i.e., f46-f49, f43-f46-f49 and f40-f43-f46-f49. Figure~\ref{fig:impacts}(b)~illustrates the impact of different layer numbers on the~average~ASuR.~As the layer number increases, ASuR mostly shows a declining trend. Compared to the single-layer setting, the average ASuR for~the~2-layer setting decreases by \todo{10.60\%}, with the 3-layer setting decreasing by \todo{10.53\%} and the 4-layer setting by \todo{7.96\%}.
The intuition~is~that the gradients arising from different layers produce conflicting influences on the overall optimization, which hinders the finding~of~the global optimum. Therefore, we use the single-layer setting.

\textbf{Impact of Neuron Number.} We configure the number of selected neurons from 10 to 150 by a step of 10. Figure~\ref{fig:impacts}(c) shows~the~impact of different neuron numbers on the average ASuR. Except~for some cases, most attacks exhibit similar performance across different numbers of selected neurons.
Considering that more neurons bring higher computation complexity, we use 10 selected neurons. 

\begin{table}[!t]
    \caption{Ablation analysis of different steps in \tool.}
    \label{tbl:ablation}
    \footnotesize
    \vspace{-10pt}
    \centering
    \begin{tabular}{c*{8}{m{0.45cm}}}
        \toprule
        Ablated Step & Bad-Nets & Blend & Trojan-NN & LC & SSBA & Input-aware & WaNet & Adap-Blend \\
        \midrule 
        Micro-Training & 63.79 & 80.82 & 53.37 & 68.63 & 76.04 & 16.09 & 47.08 & 52.28 \\
        TCDP Generation & 43.26 & 87.31 & 71.61 & 73.51 & 68.11 & 32.41 & 36.65 & 57.53\\
        Optimized BTT & 29.62 & 74.66 & 72.02 & 57.10 & 66.14 & 35.56 & 26.49 & 44.99 \\
        \midrule
        Full \tool & \textbf{64.95} & \textbf{94.72} & \textbf{86.83} & \textbf{75.55} & \textbf{76.18} & \textbf{47.23} & \textbf{51.18} & \textbf{61.25} \\
        \bottomrule
    \end{tabular}
\end{table}

\textbf{Impact of Different Steps in \tool.} Table~\ref{tbl:ablation} reports~the~average ASuR of \tool-enhanced attacks against eight defenses when each step of \tool is ablated. First, we remove the micro-training step and initialize the backdoored model randomly. \tool suffers~a significant drop in ASuR on Blend, TrojanNN and Inputaware; and the average ASuR decreases by \todo{10.30\%}. Second, we randomly select neurons from the target layer instead of selecting crucial neurons~in the TCDP generation step. The average ASuR exhibits an decrease~of \todo{10.92\%}. Third, we remove our optimization in binary-task training (BTT) step, and set a fixed weight of 1 to $\omega_1$ and $\omega_2$. The average ASuR decreases by \todo{18.90\%} across all attacks. These results indicate the importance of each step in \tool.

\textbf{Summary.} The layer depth, layer number and neuron number can affect the~survivability of \tool-enhanced attacks. Each~step in \tool contributes the performance of \tool, with our optimization in binary-task training having the most contribution.

\subsection{Explainability Analysis (RQ4)}

We employ three DNN explainability techniques, i.e., Grad-CAM~analysis, neuron activation analysis, and network feature analysis,~to~delve into the intrinsic mechanisms of \tool. These techniques~allow~us to comprehend the decision-making process of the model from different dimensions, thus explaining the performance of \tool.

\begin{figure*}[!t]
    \centering
    \includegraphics[width=.84\textwidth]{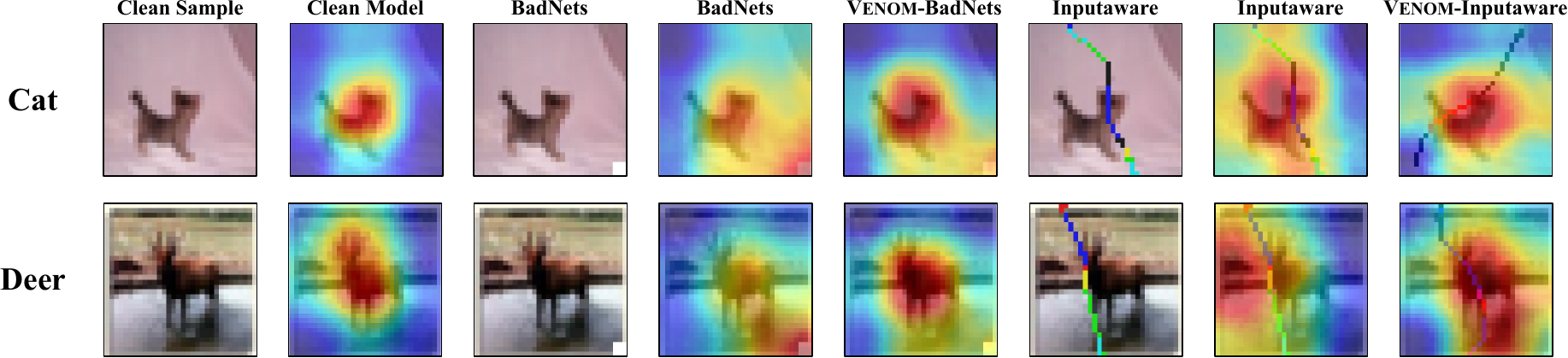}
    \vspace{-7pt}
    \caption{Grad-CAM visualization of the regions contributing to the model decision under BadNets and Inputaware. 
    }
    \label{fig:cam}
\end{figure*}

\textbf{Grad-CAM Analysis.} Gradient-weighted Class Activation Mapping (Grad-CAM)~\cite{selvaraju2017gradcam} provides an intuitive visual explanation of the model's decision-making process by generating heatmaps that reveal the key regions that the model focuses on. Grad-CAM is also frequently used in poisoned sample detection to localize trigger~regions~\cite{chou2020sentineta} without any prior knowledge of the attack.~Here,~we~employ Grad-CAM to visualize how \tool enhances BadNets and~Inputaware, given that these two attacks employ visible local triggers which can be easily visualized by Grad-CAM.

As~illustrated in Figure~\ref{fig:cam}, Grad-CAM successfully localizes the trigger regions generated by the original BadNets and Inputaware attacks. However, when these attacks are equipped with \tool, the key regions that contribute to the model's classification decision~for~poisoned images shift from the trigger to the label object~itself, hence making these attacks more stealthy and survivable.

\textbf{Neuron Activation Analysis.} Neuron activation analysis~\cite{rathore2021topoact} examines the activation patterns of individual neurons or groups~of neurons in response to different samples. It can gain insights into~how specific parts of the model contribute to the decision-making process. Here, we explore the cosine similarity of activation values~of the 10 crucial neurons between clean samples in the clean~model and poisoned samples in the backdoored model, particularly analyzing the impact of \tool on neuron activation behavior. Notice~that these 10 neurons are considered as crucial for the model's decision-making process, and if they exhibit different activation values from the clean model in the presence of backdoor, they can be the key for defenders to identify and eliminate the backdoor.

\begin{figure}[!t]
    \centering
    \includegraphics[width=.41\textwidth]{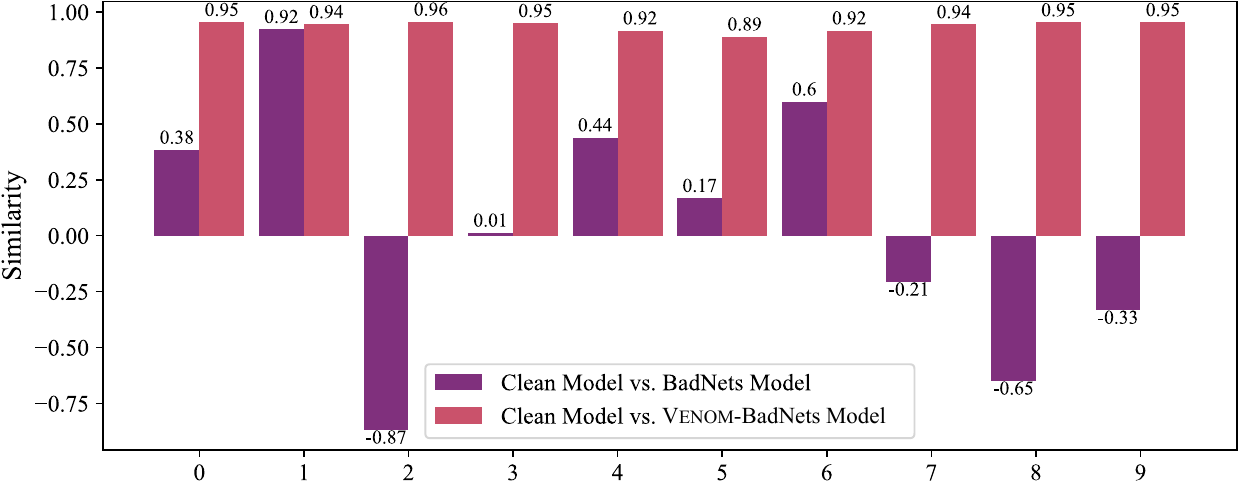}
    \vspace{-8pt}
    \caption{Similarity of crucial neuron activation between clean samples in clean model and poisoned samples in backdoored VGG19-BN model by BadNets and \tool-BadNets.}
    \label{fig:neuron_activation}
\end{figure}

As shown in Figure~\ref{fig:neuron_activation}, the backdoored model by BadNets has~a relatively low similarity with the clean model in terms of the activation values of the 10 crucial neurons (i.e., denoted~by~the $x$-axis). Differently, the backdoored model by \tool-BadNets achieves a very high similarity with the clean model. In other words, \tool adjusts the neuron activation, forcing the model to process poisoned samples using the TCDP instead of using additional neurons.~This not only allows the implanted features by the attacker~in~poisoned~samples to be processed by the model similarly to clean samples, greatly enhancing the stealthiness and effectiveness of the attack, but also makes the backdoored model closer to the clean model in its internal representation, making it more difficult to detect and defend~against.

\textbf{Network Feature Analysis.} Network feature analysis delves deeper into the model network, analyzing how the model extracts and utilizes features at different layers. By comparing the feature representations and their interactions at different layers,~we~can~understand the contribution of each layer to the decision-making~process. Here, we adopt Centered Kernel Alignment (CKA)~\cite{kornblith2019similarity}~to~quantify the similarity of representations between different layers. 

\begin{figure}[!t]
    \centering
    \subcaptionbox{\scriptsize Clean Model vs. Clean Model}{\includegraphics[width=.29\linewidth, height=2.2cm]{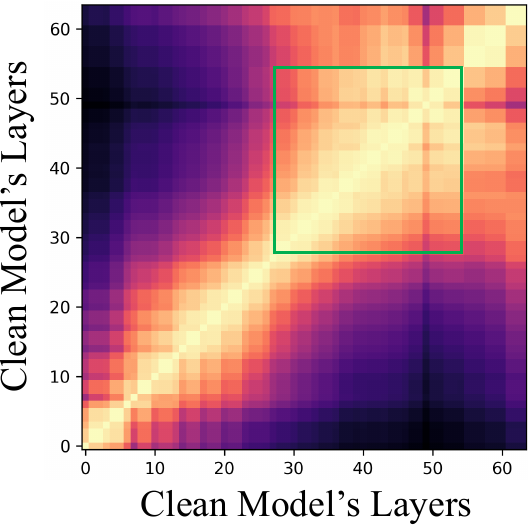}\vspace{-5pt}}\hfill
    \subcaptionbox{\scriptsize Clean Model vs. BadNets Model}{\includegraphics[width=.29\linewidth,height=2.2cm]{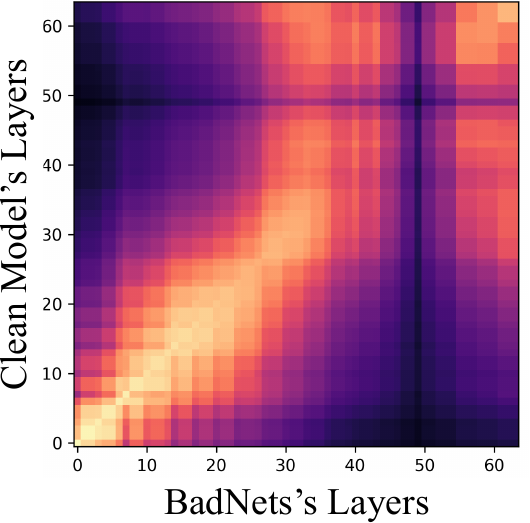}\vspace{-5pt}}\hfill
    \subcaptionbox{\scriptsize Clean Model vs. \tool-BadNets Model}{\includegraphics[width=.29\linewidth,height=2.2cm]{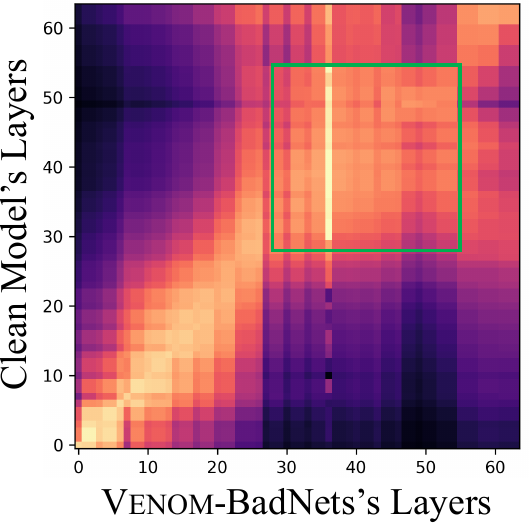}\vspace{-5pt}}\hfill
    \subcaptionbox*{}{\includegraphics[width=.08\linewidth,height=2.05cm]{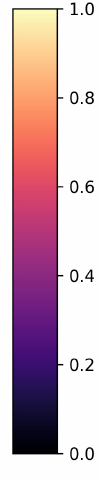}}
    \vspace{-8pt}
    \caption{Comparison of network feature analysis between clean model and backdoored PreActResNet18 model by BadNets and \tool-BadNets (with block structures highlighted).}
    \label{fig:CKA}
\end{figure}

As shown in Figure~\ref{fig:CKA}, the clean model and the backdoored~models by BadNets and \tool-BadNets exhibit some similarity in the shallow layers and the final linear layer.
However, in the shallow layers of neural networks, models tend to learn generic, low-level visual features such as edges, textures, and colors. Hence, the shallow representations of different models are usually similar. 
For the final linear layer, all models aim to map high-level features~into~probability distributions, and thus they are also quite similar. 
However,~for the deeper convolutional layers crucial for classification tasks (layers around the 45th layer), the backdoored~model by BadNets shows significant differences from the clean model, which can be exploited by defenders. 
Hence, we choose the last convolutional layer as the target layer to implement \tool, aiming to mitigate this difference.

Further, we observe the presence of \textit{block structures} (i.e., many~consecutive hidden layers that have highly similar representations \cite{nguyen2020wide}) in Figure~\ref{fig:CKA}(c), revealing similarity between the clean model~and~the backdoored model by \tool-BadNets.
Similar block structures~also appear in Figure~\ref{fig:CKA}(a). 
Compared to the original BadNets as shown~in Figure~\ref{fig:CKA}(b), \tool-BadNets successfully enhances the backdoored model to exhibit similarity to the clean model in the TCDP. 
In addition, \tool also achieves this similarity around the TCDP (i.e.,~the deeper convolutional layers) through forward and backward propagation during training, thereby enhancing the attack survivability.

\textbf{Summary.} The explainability techniques explain the rationality of the key designs in \tool as well as the performance of \tool.

\section{Discussion}

\textbf{Enhancing Poison-Only Attacks.} Our threat model assumes that the attacker possesses control over the training process, and thus \tool cannot be used to enhance poison-only attacks.~It~remains~an open problem to enhance poison-only attacks' survivability against model reconstruction-based defenses, e.g., how~to~poison the samples so that they have similar activation behaviors to benign ones.

\textbf{Against Data Distribution-Based Defenses.} Our work~is~solely focused on enhancing the survivability against model reconstruction-based defenses. However, as shown by Table~\ref{tbl:detection} in Section~\ref{sec:rq1},~our~experiments have demonstrated that Venom's enhancement is independent of the original attack's resistance against data distribution-based defenses. Therefore, the enhancements are combinable.

\textbf{Generalized to Other Tasks.} We focus on image classification tasks in this work, but \tool can be applicable to other model~architectures. Specifically, adapting the implementation for TCDP~generation in different model structures is needed. Therefore, we believe that Venom can be generalized to other tasks, such as facial~recognition and speech recognition. Our preliminary attempt on the face recognition task can be found in Appendix~\ref{Appendix:other_tasks}.

\textbf{Potential Defenses.} Inspired by Differentially Private SGD~\cite{abadi2016deep} which adds noise to the gradient during the training process without affecting the final performance of the model and meanwhile~protecting the privacy of the training data, one potential defense against \tool is to apply a certain range of perturbations to model weights. As we assume that backdoor-related neurons are far less than the benign classification-related neurons, the backdoor is more sensitive to changes in model weights. Thus,~the~decision boundary~of~the perturbed backdoored model is unlikely to undergo significant shifts, thus preserving a high BA while exhibiting significant decrease in ASR. Nonetheless, the challenge remains formidable when it comes to effectively introducing perturbations to the model's weights and precisely determining the optimal range for such perturbations.

\section{Conclusions}

We have proposed \tool, the first generic approach to enhance the survivability of existing backdoor attacks against existing model reconstruction-based backdoor defenses. 
Our extensive evaluation has demonstrated the performance of tool.
In future, we plan~to~evaluate \tool on more tasks and develop defenses against \tool.

\bibliographystyle{ACM-Reference-Format}
\bibliography{reference}

\appendix
\section{Datasets and Models}\label{Appendix:Datasets and Models}

Table~\ref{tbl:Datasets} reports detailed information of the three datasets,~including the number of classes, the size of the training and testing datasets, and the size of the image. They are commonly employed~in~the~evaluation of backdoor attacks~\cite{gu2019badnets,chen2017targeted,turner2019label,li2021invisible}.

Table~\ref{tbl:Models} provides detailed information of the DNN models.
VGG19-BN and PreActResNet18 are implemented following the default~configuration in Backdoorbench~\cite{wu2022backdoorbench}.
The benign accuracy of the four classification tasks are also presented in the last column,~which~serves as baselines for comparing the utility of backdoored models.

\section{Attack Setup}\label{Appendix:Attack Setting}

We set the attack target to label 0 for all tasks. However, LC fails to inject backdoor into the model using the GTSRB dataset.~The~reason is that LC attack only poisons the target class, but the GTSRB~dataset has only 210 images for class 0, which is insufficient for backdoor injection. Therefore, we set the attack target to label 1 for LC and \tool-LC on the GTSRB dataset.
As Adap-Blend partially stems from a low poison rate, we follow its original configuration to set its poison rate to 0.003 and its attack target on GTSRB to 2. Besides, the total epochs for most tasks are 80 epochs, but 60 epochs are already sufficient for the GTSRB dataset.

With respect to the specific setting of \tool, the micro-training step takes 5\% of the whole epochs, which is 4 epochs for CIFAR-10 and CIFAR-100 and 3 epochs for GTSRB. In the TCDP generation step, we uniformly select the last convolutional layer as the target layer to generate the target crucial decision path, i.e., \textit{features.49} for VGG19-BN and \textit{layer4.1.conv2} for PreActResNet18.

In the binary-task training step, we set $s$ to 3 epochs. The value of $s$ is related to the difficulty of the task. In our scenario, $\mathcal{L}_2$ drops quickly below 0.5 in the first few epochs, and thus we should set $s$ to a small value to slow down the decline of $\mathcal{L}_2$.
After $s$ epochs, $\mathcal{L}_2$ updates at a relatively slow rate, waiting for $\mathcal{L}_1$ to catch up and ultimately reaching a balance point.

\begin{table}[!t]
    \caption{Details of datasets.}
    \label{tbl:Datasets}
    \footnotesize
    \vspace{-10pt}
    \centering
    \begin{tabular}{cm{0.9cm}m{1.8cm}m{1.5cm}}
        \toprule
        Datasets & Classes & Size (train/test) & Image Size  \\ 
        \midrule 
        CIFAR-10~\cite{krizhevsky2009learning} & 10 & 50,000/10,000 & 32$\times$32 \\
        CIFAR-100~\cite{krizhevsky2009learning} & 100 & 50,000/10,000 & 32$\times$32 \\
        GTSRB~\cite{6706807} & 43 & 39,209/12,630 & 32$\times$32 \\
        \bottomrule
    \end{tabular}
\end{table}

\begin{table}[!t]
    \caption{Models and their parameters.}
    \label{tbl:Models}
    \footnotesize
    \vspace{-10pt}
    \centering
    \begin{tabular}{cm{1.3cm}m{1.15cm}m{1.15cm}}
        \toprule
        Model & Training Dataset & Parameters & Accuracy \\
        \midrule 
        \multirow{3}{*}{VGG19-BN~\cite{simonyan2015very}} & CIFAR-10 & 139.62M & 92.23\% \\
        & CIFAR-100 & 139.99M & 65.47\% \\
        & GTSRB & 139.76M & 98.12\% \\
\midrule
        \multirow{1}{*}{PreActResNet18~\cite{he2016identity}} & CIFAR-10 & 11.17M & 93.83\% \\
        \bottomrule 
    \end{tabular}
\end{table}

\begin{table*}[!t]
    \caption{Attack survivability comparison on VGG19-BN using CIFAR-100.}
    \label{tbl:vgg19 on cifar100}
    \centering
    \vspace{-8pt}
    \footnotesize
    \begin{tabular}{cc *{18}{p{0.51cm}}}
        \toprule
        \multirow{2}{*}{Defense} & \multirow{2}{*}{Metric} & 
            \multicolumn{2}{c}{BadNets} &
            \multicolumn{2}{c}{Blend} & 
            \multicolumn{2}{c}{TrojanNN} &
            \multicolumn{2}{c}{LC} & 
            \multicolumn{2}{c}{SSBA} & 
            \multicolumn{2}{c}{Inputaware} & 
            \multicolumn{2}{c}{WaNet} & 
            \multicolumn{2}{c}{Adap-Blend} & 
            \multicolumn{2}{c}{{Avg}} \\
            \cmidrule(lr){3-4}\cmidrule(lr){5-6}\cmidrule(lr){7-8}\cmidrule(lr){9-10}\cmidrule(lr){11-12}\cmidrule(lr){13-14}\cmidrule(lr){15-16}\cmidrule(lr){17-18}\cmidrule(lr){19-20}
            & & O. & V. & O. & V. &  O. & V. &  O. & V. &  O. & V. &  O. & V. &  O. & V. &  O. & V. & {O.} & {V.} \\ 
        \midrule  
        \multirow{3}{*}{FT} & BA & 59.06 & 60.17 & 61.83 & 64.81 & 61.22 & 65.58 & 63.23 & 66.54 & 60.40 & 62.78 & 64.06 & 62.70 & 64.46 & 62.35 & 63.62 & 68.11 & 62.23 & \textbf{64.13} \\
        & ASR & 0.09 & 43.10 & 89.71 & 99.14 & 0.75 & 100.00 & 7.81 & 58.63 & 37.44 & 83.91 & 1.58 & 57.92 & 8.12 & 30.35 & 18.11 & 58.59 & 20.45 & \textbf{66.45} \\
        & ASuR & 0.91 & 42.09 & 85.99 & 95.29 & 1.40 & 96.11 & 8.28 & 56.85 & 36.26 & 80.81 & 3.03 & 56.44 & 9.48 & 30.66 & 17.82 & 56.81 & 20.40 & \textbf{64.38} \\
        \midrule 
        \multirow{3}{*}{I-BAU} & BA & 56.27 & 57.77 & 57.41 & 62.76 & 59.29 & 63.37 & 58.38 & 63.04 & 56.24 & 61.38 & 61.00 & 62.70 & 60.82 & 59.94 & 62.79 & 65.84 & 59.02 & \textbf{62.10} \\
        & ASR & 0.10 & 58.25 & 43.08 & 98.41 & 0.82 & 99.84 & 23.67 & 66.25 & 87.83 & 89.00 & 61.21 & 57.92 & 0.63 & 7.83 & 9.58 & 39.36 & 28.37 & \textbf{64.61} \\
        & ASuR & 0.64 & 56.25 & 41.21 & 94.37 & 1.26 & 95.71 & 22.80 & 63.69 & 83.67 & 85.50 & 59.38 & 56.44 & 2.03 & 9.06 & 9.62 & 38.27 & 27.58 & \textbf{62.41} \\
        \midrule 
        \multirow{3}{*}{NAD} & BA & 57.84 & 59.93 & 59.44 & 64.57 & 59.00 & 64.53 & 63.63 & 66.42 & 57.80 & 62.02 & 63.43 & 62.22 & 63.88 & 61.85 & 62.53 & 67.71 & 60.94 & \textbf{63.66} \\
        & ASR & 0.12 & 22.77 & 56.64 & 99.18 & 4.15 & 100.00 & 22.69 & 57.09 & 75.05 & 89.53 & 0.68 & 58.51 & 1.87 & 29.43 & 16.79 & 67.78 & 22.25 & \textbf{65.54} \\
        & ASuR & 0.82 & 22.76 & 54.31 & 95.30 & 4.39 & 95.99 & 22.46 & 55.38 & 71.70 & 86.07 & 2.11 & 56.95 & 3.49 & 29.74 & 16.44 & 65.50 & 21.97 & \textbf{63.46} \\
        \midrule 
        \multirow{3}{*}{NC} & BA & 59.19 & 59.86 & 60.75 & 63.87 & 61.18 & 65.01 & 61.53 & 66.49 & 60.98 & 62.33 & 63.62 & 61.90 & 63.55 & 62.01 & 68.26 & 68.53 & 62.38 & \textbf{63.75} \\
        & ASR & 0.28 & 24.67 & 0.01 & 99.04 & 1.01 & 99.98 & 4.40 & 55.76 & 57.01 & 95.39 & 0.96 & 43.75 & 0.33 & 16.59 & 51.81 & 71.09 & 14.48 & \textbf{63.28} \\
        & ASuR & 1.11 & 24.56 & 0.66 & 95.09 & 1.65 & 96.03 & 4.85 & 54.12 & 54.91 & 91.67 & 2.40 & 42.90 & 1.99 & 17.56 & 50.40 & 68.74 & 14.75 & \textbf{61.33} \\
        \midrule 
        \multirow{3}{*}{FP} & BA & 61.05 & 60.19 & 63.37 & 64.57 & 63.58 & 65.37 & 64.84 & 66.54 & 62.88 & 61.90 & 63.67 & 61.83 & 64.63 & 62.63 & 66.59 & 68.78 & 63.83 & \textbf{63.98} \\
        & ASR & 0.57 & 29.24 & 90.88 & 98.58 & 99.35 & 99.67 & 20.02 & 55.66 & 77.24 & 87.51 & 0.19 & 47.62 & 11.05 & 9.19 & 27.14 & 55.29 & 40.80 & \textbf{60.34} \\
        & ASuR & 1.57 & 28.93 & 87.28 & 94.73 & 95.34 & 95.77 & 20.06 & 54.03 & 74.34 & 84.14 & 1.67 & 46.57 & 12.27 & 10.58 & 26.76 & 53.76 & 39.91 & \textbf{58.56} \\
        \midrule 
        \multirow{3}{*}{BNP} & BA & 60.35 & 58.40 & 64.77 & 64.65 & 63.90 & 65.77 & 64.94 & 66.28 & 63.38 & 58.41 & 58.82 & 58.00 & 50.60 & 15.21 & 67.14 & 64.51 & \textbf{61.74} & 56.40 \\
        & ASR & 89.02 & 90.25 & 98.68 & 98.98 & 99.75 & 99.95 & 19.21 & 59.30 & 95.24 & 94.36 & 90.38 & 86.48 & 84.27 & 95.94 & 51.72 & 59.85 & 78.53 & \textbf{85.64} \\
        & ASuR & 85.53 & 86.72 & 94.84 & 95.12 & 95.75 & 96.08 & 19.30 & 57.46 & 91.49 & 90.28 & 86.88 & 83.12 & 80.55 & 79.30 & 50.18 & 57.58 & 75.56 & \textbf{80.71} \\
        \midrule 
        \multirow{3}{*}{CLP} & BA & 56.91 & 25.67 & 59.16 & 62.49 & 56.83 & 61.51 & 62.87 & 64.57 & 57.42 & 46.82 & 49.46 & 33.40 & 58.95 & 50.72 & 53.27 & 64.60 & \textbf{56.86} & 51.22 \\
        & ASR & 85.17 & 23.75 & 93.83 & 98.59 & 62.00 & 98.09 & 6.18 & 16.05 & 96.86 & 81.18 & 43.72 & 17.89 & 11.31 & 60.44 & 17.95 & 55.62 & 52.13 & \textbf{56.45} \\
        & ASuR & 81.52 & 35.30 & 89.61 & 94.51 & 59.10 & 93.83 & 6.69 & 16.18 & 92.38 & 45.85 & 41.65 & 24.15 & 12.00 & 58.26 & 13.44 & 53.57 & 49.55 & \textbf{52.71} \\
        \midrule 
        \multirow{3}{*}{NPD} & BA & 60.51 & 58.40 & 62.01 & 62.51 & 62.12 & 63.64 & 63.60 & 64.49 & 61.73 & 61.43 & 60.75 & 59.67 & 60.34 & 58.84 & 66.18 & 67.48 & \textbf{62.15} & 62.06 \\
        & ASR & 0.01 & 54.51 & 9.43 & 92.74 & 2.58 & 99.73 & 15.63 & 44.22 & 0.86 & 73.48 & 0.12 & 1.92 & 47.72 & 0.36 & 29.78 & 56.04 & 13.27 & \textbf{52.88} \\
        & ASuR & 0.99 & 52.76 & 9.75 & 88.95 & 3.24 & 95.63 & 15.75 & 42.93 & 1.65 & 70.76 & 1.32 & 2.95 & 46.72 & 1.87 & 29.22 & 54.32 & 13.58 & \textbf{51.27} \\
        \midrule 
        \multirow{3}{*}{Avg} & BA & \textbf{58.90} & 55.05 & 61.09 & \textbf{63.78} & 60.89 & \textbf{64.35} & 62.88 & \textbf{65.55} & \textbf{60.10} & 59.63 & \textbf{60.60} & 57.80 & \textbf{60.90} & 54.19 & 63.80 & \textbf{66.94} & \textbf{61.15} & 60.91 \\
        & ASR & 21.92 & \textbf{43.32} & 60.28 & \textbf{98.08} & 33.80 & \textbf{99.66} & 14.95 & \textbf{51.62} & 65.94 & \textbf{86.79} & 24.86 & \textbf{46.50} & 20.66 & \textbf{31.27} & 27.86 & \textbf{57.95} & 33.78 & \textbf{64.40} \\
        & ASuR & 21.64 & \textbf{43.67} & 57.96 & \textbf{94.17} & 32.77 & \textbf{95.64} & 15.02 & \textbf{50.08} & 63.30 & \textbf{79.39} & 24.81 & \textbf{46.19} & 21.07 & \textbf{29.63} & 26.73 & \textbf{56.07} & 32.91 & \textbf{61.85} \\        
        \bottomrule
    \end{tabular}
\end{table*}

\begin{table*}[!t]
    \caption{Attack survivability comparison on VGG19-BN using GTSRB.}
    \label{tbl:vgg19 on gtsrb}
    \centering
    \vspace{-8pt}
    \footnotesize
    \begin{tabular}{cc *{18}{p{0.51cm}}}
        \toprule
        \multirow{2}{*}{Defense} & \multirow{2}{*}{Metric} & 
            \multicolumn{2}{c}{BadNets} &
            \multicolumn{2}{c}{Blend} & 
            \multicolumn{2}{c}{TrojanNN} &
            \multicolumn{2}{c}{LC} & 
            \multicolumn{2}{c}{SSBA} & 
            \multicolumn{2}{c}{Inputaware} & 
            \multicolumn{2}{c}{WaNet} & 
            \multicolumn{2}{c}{Adap-Blend} & 
            \multicolumn{2}{c}{{Avg}} \\
            \cmidrule(lr){3-4}\cmidrule(lr){5-6}\cmidrule(lr){7-8}\cmidrule(lr){9-10}\cmidrule(lr){11-12}\cmidrule(lr){13-14}\cmidrule(lr){15-16}\cmidrule(lr){17-18}\cmidrule(lr){19-20}
            & & O. & V. & O. & V. &  O. & V. &  O. & V. &  O. & V. &  O. & V. &  O. & V. &  O. & V. & {O.} & {V.} \\ 
        \midrule  
        \multirow{3}{*}{FT} & BA & 97.74 & 97.85 & 97.78 & 98.23 & 98.34 & 98.31 & 97.81 & 98.12 & 97.84 & 97.50 & 97.80 & 97.27 & 98.34 & 98.42 & 95.83 & 96.77 & 97.69 & \textbf{97.81} \\
        & ASR & 8.80 & 89.84 & 99.96 & 99.99 & 17.30 & 100.00 & 69.37 & 57.69 & 99.60 & 99.71 & 18.86 & 76.25 & 1.20 & 26.98 & 28.37 & 67.23 & 42.93 & \textbf{77.21} \\
        & ASuR & 12.46 & 89.45 & 99.05 & 99.23 & 20.74 & 99.27 & 70.00 & 59.04 & 98.74 & 98.73 & 22.19 & 76.44 & 5.59 & 30.11 & 30.32 & 67.67 & 44.89 & \textbf{77.49} \\
        \midrule 
        \multirow{3}{*}{I-BAU} & BA & 96.85 & 97.34 & 96.48 & 97.32 & 24.23 & 96.85 & 5.29 & 97.59 & 96.61 & 96.71 & 95.73 & 95.87 & 97.19 & 97.73 & 85.65 & 94.83 & 74.75 & \textbf{96.78} \\
        & ASR & 0.02 & 0.00 & 97.64 & 97.75 & 0.00 & 24.89 & 0.00 & 41.74 & 78.40 & 81.49 & 0.60 & 0.80 & 1.65 & 0.53 & 25.80 & 44.18 & 25.51 & \textbf{36.42} \\
        & ASuR & 3.76 & 3.89 & 96.31 & 96.71 & 36.25 & 27.29 & 46.99 & 43.68 & 78.10 & 81.11 & 4.15 & 4.24 & 5.64 & 4.75 & 13.79 & 45.06 & 35.62 & \textbf{38.34} \\
        \midrule 
        \multirow{3}{*}{NAD} & BA & 97.56 & 98.04 & 97.73 & 98.27 & 98.14 & 98.12 & 97.79 & 98.00 & 97.70 & 97.44 & 97.62 & 97.20 & 98.20 & 98.47 & 95.84 & 96.54 & 97.57 & \textbf{97.76} \\
        & ASR & 60.06 & 91.58 & 99.98 & 99.97 & 96.01 & 100.00 & 69.34 & 61.70 & 99.74 & 99.24 & 13.62 & 81.34 & 0.23 & 30.29 & 34.75 & 70.63 & 59.22 & \textbf{79.34} \\
        & ASuR & 61.08 & 91.19 & 99.05 & 99.23 & 95.43 & 99.19 & 69.97 & 62.81 & 98.81 & 98.26 & 17.15 & 81.25 & 4.63 & 33.27 & 36.39 & 70.82 & 60.31 & \textbf{79.50} \\
        \midrule 
        \multirow{3}{*}{NC} & BA & 95.99 & 96.65 & 95.38 & 96.39 & 94.49 & 95.08 & 94.09 & 96.17 & 95.80 & 95.74 & 96.19 & 96.29 & 96.27 & 97.22 & 74.47 & 96.48 & 92.84 & \textbf{96.25} \\
        & ASR & 0.01 & 95.06 & 67.68 & 81.23 & 32.08 & 100.00 & 25.21 & 44.51 & 46.68 & 91.75 & 65.23 & 0.21 & 0.29 & 10.86 & 49.77 & 71.11 & 35.87 & \textbf{61.84} \\
        & ASuR & 3.41 & 93.91 & 67.39 & 80.62 & 33.16 & 97.89 & 26.52 & 45.73 & 47.63 & 90.47 & 65.70 & 3.84 & 4.05 & 14.40 & 32.19 & 71.25 & 35.01 & \textbf{62.26} \\
        \midrule 
        \multirow{3}{*}{FP} & BA & 97.81 & 98.07 & 97.95 & 98.31 & 98.41 & 98.10 & 97.74 & 98.30 & 98.09 & 97.50 & 97.55 & 97.24 & 98.24 & 98.57 & 95.74 & 96.71 & 97.69 & \textbf{97.85} \\
        & ASR & 0.00 & 91.92 & 99.89 & 99.98 & 100.00 & 100.00 & 65.37 & 59.05 & 98.81 & 99.15 & 2.27 & 19.04 & 0.18 & 4.60 & 31.97 & 54.60 & 49.81 & \textbf{66.04} \\
        & ASuR & 4.12 & 91.52 & 99.05 & 99.26 & 99.33 & 99.18 & 66.17 & 60.41 & 98.09 & 98.20 & 6.34 & 22.08 & 4.59 & 8.90 & 33.71 & 55.65 & 51.42 & \textbf{66.90} \\
        \midrule 
        \multirow{3}{*}{BNP} & BA & 97.43 & 97.95 & 97.83 & 98.41 & 98.17 & 98.29 & 97.72 & 97.55 & 97.59 & 97.48 & 95.67 & 94.43 & 13.39 & 89.09 & 75.24 & 96.46 & 84.13 & \textbf{96.21} \\
        & ASR & 94.14 & 95.39 & 99.94 & 99.98 & 68.27 & 100.00 & 63.89 & 64.31 & 99.56 & 99.07 & 61.92 & 34.42 & 99.86 & 97.74 & 43.10 & 67.21 & 78.83 & \textbf{82.27} \\
        & ASuR & 93.41 & 94.77 & 99.05 & 99.30 & 69.09 & 99.27 & 64.76 & 65.10 & 98.60 & 98.12 & 62.38 & 35.66 & 92.03 & 94.25 & 28.41 & 67.54 & 75.97 & \textbf{81.75} \\
        \midrule 
        \multirow{3}{*}{CLP} & BA & 97.54 & 97.74 & 97.33 & 98.19 & 97.82 & 98.22 & 97.56 & 96.99 & 97.68 & 97.35 & 96.30 & 95.96 & 76.45 & 35.99 & 74.34 & 96.52 & \textbf{91.88} & 89.62 \\
        & ASR & 15.76 & 2.38 & 98.58 & 99.81 & 1.04 & 57.18 & 13.59 & 26.67 & 99.56 & 99.01 & 17.49 & 66.75 & 94.89 & 90.58 & 50.73 & 66.94 & 48.96 & \textbf{63.66} \\
        & ASuR & 18.99 & 6.32 & 97.55 & 99.04 & 5.07 & 58.56 & 16.91 & 29.12 & 98.64 & 98.01 & 20.39 & 66.93 & 52.36 & 74.08 & 32.74 & 67.31 & 42.83 & \textbf{62.42} \\
        \midrule 
        \multirow{3}{*}{NPD} & BA & 97.19 & 96.64 & 97.40 & 97.14 & 95.41 & 96.69 & 96.68 & 97.18 & 96.98 & 96.94 & 96.55 & 95.34 & 97.02 & 96.37 & 84.48 & 95.73 & 95.21 & \textbf{96.50} \\
        & ASR & 0.00 & 0.00 & 9.27 & 10.21 & 0.05 & 99.96 & 29.39 & 49.56 & 4.12 & 49.68 & 0.00 & 36.27 & 19.69 & 45.58 & 38.84 & 41.12 & 12.67 & \textbf{41.55} \\
        & ASuR & 3.88 & 3.60 & 12.74 & 13.47 & 3.12 & 98.54 & 31.56 & 50.94 & 7.68 & 50.98 & 3.85 & 37.75 & 22.73 & 47.10 & 20.99 & 42.48 & 13.32 & \textbf{43.11} \\
        \midrule 
        \multirow{3}{*}{Avg} & BA & 97.26 & \textbf{97.53} & 97.23 & \textbf{97.78} & 88.13 & \textbf{97.46} & 85.59 & \textbf{97.49} & \textbf{97.29} & 97.08 & \textbf{96.68} & 96.20 & 84.39 & \textbf{88.98} & 85.20 & \textbf{96.25} & 91.47 & \textbf{96.10} \\
        & ASR & 22.35 & \textbf{58.27} & 84.12 & \textbf{86.12} & 39.34 & \textbf{85.25} & 42.02 & \textbf{50.65} & 78.31 & \textbf{89.89} & 22.50 & \textbf{39.38} & 27.25 & \textbf{38.39} & 37.92 & \textbf{60.38} & 44.23 & \textbf{63.54} \\
        & ASuR & 25.14 & \textbf{59.33} & 83.77 & \textbf{85.86} & 45.27 & \textbf{84.90} & 49.11 & \textbf{52.10} & 78.29 & \textbf{89.24} & 25.27 & \textbf{41.02} & 23.95 & \textbf{38.36} & 28.57 & \textbf{60.97} & 44.92 & \textbf{63.97} \\             
        \bottomrule
    \end{tabular}
\end{table*}

\begin{table*}[!t]
    \caption{Attack survivability comparison on PreActResNet18 using CIFAR-10.}
    \label{tbl:resnet on cifar10}
    \centering
    \vspace{-8pt}
    \footnotesize
    \begin{tabular}{cc *{18}{p{0.51cm}}}
        \toprule
        \multirow{2}{*}{Defense} & \multirow{2}{*}{Metric} & 
            \multicolumn{2}{c}{BadNets} &
            \multicolumn{2}{c}{Blend} & 
            \multicolumn{2}{c}{TrojanNN} &
            \multicolumn{2}{c}{LC} & 
            \multicolumn{2}{c}{SSBA} & 
            \multicolumn{2}{c}{Inputaware} & 
            \multicolumn{2}{c}{WaNet} & 
            \multicolumn{2}{c}{Adap-Blend} & 
            \multicolumn{2}{c}{{Avg}} \\
            \cmidrule(lr){3-4}\cmidrule(lr){5-6}\cmidrule(lr){7-8}\cmidrule(lr){9-10}\cmidrule(lr){11-12}\cmidrule(lr){13-14}\cmidrule(lr){15-16}\cmidrule(lr){17-18}\cmidrule(lr){19-20}
            & & O. & V. & O. & V. &  O. & V. &  O. & V. &  O. & V. &  O. & V. &  O. & V. &  O. & V. & {O.} & {V.} \\ 
        \midrule
        \multirow{3}{*}{FT} & BA & 90.37 & 89.75 & 93.05 & 92.74 & 92.28 & 93.40 & 91.26 & 92.26 & 92.27 & 93.06 & 93.02 & 92.98 & 93.16 & 92.23 & 93.47 & 93.42 & 92.36 & \textbf{92.48} \\
        & ASR & 1.73 & 75.10 & 95.04 & 98.52 & 3.82 & 3.19 & 35.56 & 49.57 & 80.64 & 94.52 & 73.41 & 95.20 & 15.32 & 35.13 & 71.18 & 76.82 & 47.09 & \textbf{66.01} \\
        & ASuR & 4.05 & 73.79 & 93.16 & 96.44 & 6.27 & 6.00 & 37.04 & 50.53 & 79.33 & 92.71 & 72.89 & 93.55 & 17.83 & 36.43 & 70.62 & 76.03 & 47.65 & \textbf{65.69} \\
        \midrule 
        \multirow{3}{*}{I-BAU} & BA & 88.08 & 88.42 & 89.34 & 91.44 & 88.57 & 90.98 & 88.26 & 88.96 & 89.78 & 89.85 & 91.27 & 91.72 & 91.86 & 91.32 & 88.75 & 91.21 & 89.49 & \textbf{90.49} \\
        & ASR & 2.19 & 1.76 & 8.63 & 53.30 & 1.86 & 3.54 & 11.61 & 21.04 & 6.21 & 28.08 & 0.17 & 31.24 & 2.01 & 62.50 & 16.40 & 47.39 & 6.13 & \textbf{31.11} \\
        & ASuR & 3.87 & 3.79 & 9.93 & 53.09 & 3.27 & 5.59 & 13.69 & 22.76 & 7.88 & 28.63 & 2.84 & 32.45 & 4.86 & 62.20 & 17.13 & 47.42 & 7.93 & \textbf{31.99} \\
        \midrule 
        \multirow{3}{*}{NAD} & BA & 88.91 & 89.11 & 91.75 & 92.53 & 92.02 & 93.09 & 90.29 & 92.16 & 91.85 & 92.88 & 92.88 & 92.83 & 92.94 & 92.19 & 93.39 & 93.07 & 91.75 & \textbf{92.23} \\
        & ASR & 1.21 & 49.20 & 76.56 & 97.52 & 2.26 & 3.07 & 20.44 & 43.57 & 85.49 & 92.56 & 10.29 & 97.14 & 8.46 & 45.44 & 66.49 & 74.82 & 33.90 & \textbf{62.91} \\
        & ASuR & 3.17 & 49.03 & 75.20 & 95.42 & 4.70 & 5.79 & 22.48 & 44.81 & 83.81 & 90.80 & 12.88 & 95.36 & 11.26 & 46.21 & 66.14 & 74.03 & 34.95 & \textbf{62.68} \\
        \midrule 
        \multirow{3}{*}{NC} & BA & 89.63 & 89.35 & 92.33 & 92.84 & 93.67 & 93.76 & 89.78 & 85.18 & 92.38 & 93.09 & 92.82 & 91.15 & 92.28 & 92.18 & 93.68 & 93.22 & \textbf{92.07} & 91.35 \\
        & ASR & 1.60 & 45.84 & 95.67 & 91.94 & 100.00 & 99.99 & 6.39 & 92.14 & 77.18 & 93.42 & 23.08 & 95.61 & 3.07 & 19.20 & 74.63 & 73.01 & 47.70 & \textbf{76.39} \\
        & ASuR & 3.73 & 45.89 & 93.54 & 90.22 & 98.06 & 98.07 & 9.03 & 89.55 & 76.07 & 91.68 & 25.02 & 93.45 & 5.97 & 21.28 & 73.96 & 72.35 & 48.17 & \textbf{75.31} \\
        \midrule 
        \multirow{3}{*}{FP} & BA & 91.50 & 89.56 & 92.72 & 92.56 & 92.65 & 93.01 & 91.06 & 92.10 & 92.06 & 92.89 & 93.18 & 93.39 & 92.99 & 92.32 & 92.60 & 92.88 & \textbf{92.35} & 92.34 \\
        & ASR & 0.90 & 44.20 & 25.87 & 65.10 & 64.51 & 97.64 & 55.02 & 65.28 & 14.86 & 51.08 & 10.36 & 68.72 & 1.86 & 3.64 & 37.12 & 42.71 & 26.31 & \textbf{54.80} \\
        & ASuR & 3.57 & 44.39 & 27.34 & 64.63 & 64.03 & 95.61 & 55.48 & 65.42 & 16.77 & 51.39 & 13.03 & 68.51 & 5.00 & 6.53 & 38.00 & 43.47 & 27.90 & \textbf{54.99} \\
        \midrule 
        \multirow{3}{*}{BNP} & BA & 91.26 & 89.92 & 92.23 & 92.87 & 93.43 & 93.62 & 84.75 & 85.17 & 91.75 & 93.22 & 90.99 & 89.81 & 82.78 & 63.82 & 93.28 & 93.14 & \textbf{90.06} & 87.70 \\
        & ASR & 94.26 & 95.83 & 99.73 & 99.89 & 100.00 & 99.99 & 94.49 & 92.62 & 97.72 & 97.87 & 2.11 & 2.21 & 25.23 & 99.08 & 75.37 & 75.83 & 73.61 & \textbf{82.92} \\
        & ASuR & 92.20 & 93.53 & 97.36 & 97.78 & 97.99 & 98.03 & 91.72 & 90.00 & 95.40 & 95.94 & 4.61 & 4.36 & 24.64 & 59.68 & 74.54 & 75.01 & 72.31 & \textbf{76.79} \\
        \midrule 
        \multirow{3}{*}{CLP} & BA & 90.76 & 60.18 & 89.68 & 91.29 & 91.68 & 93.46 & 84.03 & 84.16 & 92.19 & 92.57 & 90.31 & 89.98 & 87.05 & 84.71 & 92.13 & 88.14 & \textbf{89.73} & 85.56 \\
        & ASR & 14.59 & 8.02 & 99.21 & 98.50 & 98.24 & 99.99 & 2.01 & 10.41 & 98.00 & 96.21 & 3.03 & 38.60 & 13.82 & 25.24 & 67.22 & 22.08 & 49.51 & \textbf{49.88} \\
        & ASuR & 16.38 & 16.37 & 96.09 & 95.99 & 95.78 & 97.98 & 3.72 & 11.70 & 95.79 & 94.17 & 5.30 & 38.98 & 14.87 & 25.15 & 66.45 & 22.47 & 49.30 & \textbf{50.35} \\
        \midrule 
        \multirow{3}{*}{NPD} & BA & 89.46 & 85.91 & 91.34 & 91.32 & 91.10 & 91.67 & 90.42 & 90.83 & 90.52 & 91.20 & 87.82 & 90.57 & 90.09 & 88.96 & 91.29 & 91.28 & \textbf{90.25} & 90.22 \\
        & ASR & 0.26 & 0.74 & 12.77 & 16.81 & 34.89 & 7.69 & 14.01 & 32.80 & 5.08 & 13.67 & 1.41 & 0.03 & 0.42 & 1.47 & 6.40 & 42.66 & 9.41 & \textbf{14.48} \\
        & ASuR & 2.41 & 2.19 & 14.48 & 18.39 & 35.42 & 9.74 & 16.40 & 34.31 & 7.03 & 15.35 & 3.10 & 2.49 & 2.91 & 3.63 & 8.41 & 42.95 & 11.27 & \textbf{16.13} \\
        \midrule 
        \multirow{3}{*}{Avg} & BA & \textbf{90.00} & 85.27 & 91.55 & \textbf{92.20} & 91.93 & \textbf{92.87} & 88.73 & \textbf{88.85} & 91.60 & \textbf{92.34} & 91.54 & \textbf{91.55} & \textbf{90.39} & 87.22 & \textbf{92.32} & 92.04 & \textbf{91.01} & 90.29 \\
        & ASR & 14.59 & \textbf{40.09} & 64.19 & \textbf{77.70} & 50.70 & \textbf{51.89} & 29.94 & \textbf{50.93} & 58.15 & \textbf{70.93} & 15.48 & \textbf{53.59} & 8.77 & \textbf{36.46} & 51.85 & \textbf{56.91} & 36.71 & \textbf{54.81} \\
        & ASuR & 16.17 & \textbf{41.12} & 63.39 & \textbf{76.49} & 50.69 & \textbf{52.10} & 31.20 & \textbf{51.13} & 57.76 & \textbf{70.08} & 17.46 & \textbf{53.64} & 10.92 & \textbf{32.64} & 51.91 & \textbf{56.72} & 37.44 & \textbf{54.24} \\              
        \bottomrule
    \end{tabular}
\end{table*}

\section{Results on the Other Three Tasks}\label{Appendix:Results of Other Datasets and Models}

Table~\ref{tbl:vgg19 on cifar100} reports the attack survivability comparison on VGG19-BN using CIFAR-100. Compared to the original attacks, \tool~helps~to increase ASR from \todo{33.78\%} to \todo{64.40\%} and ASuR from \todo{32.91\%}~to~\todo{61.85\%} on average. Specifically, the survivability enhancement on Trojan-NN is particularly significant, with an average increase~of~\todo{62.87\%} in ASuR, whereas in the case of WaNet, the enhancement is comparatively limited, with an average improvement of \todo{8.56\%} in ASuR. From the perspective of defenses, \tool generally enhances the resistance against various defenses. The resistance against NC shows the highest~improvement, with an average increase of \todo{46.58\%} in ASuR. The resistance against NPD, FT, I-BAU and NAD all demonstrate a remarkable enhancement around \todo{40\%} in ASuR.

Table~\ref{tbl:vgg19 on gtsrb} provides the attack survivability comparison on VGG19-BN using GTSRB. Overall, \tool helps to increase the average~ASR from \todo{44.23\%} to \todo{63.54\%} and ASuR from \todo{44.92\%} to \todo{63.97\%}. Specifically, the survivability enhancement on BadNets and TrojanNN~is significant, with an average improvement of \todo{34.19\%} and \todo{39.63\%}~in~ASuR respectively. From the perspective of defenses, \tool generally~enhances the resistance against various defenses. 
The resistance against FP shows the highest improvement, with an average increase of \todo{32.60\%} in ASuR, whereas only Adap-Blend exhibits significantly enhanced resistance against I-BAU.

Table~\ref{tbl:resnet on cifar10} provides the attack survivability comparison on PreActResNet18 using CIFAR-10. Generally, \tool helps to improve the average ASR from \todo{36.71\%} to \todo{54.81\%} and ASuR from \todo{37.44\%} to \todo{54.24\%}. Specifically, the survivability enhancement on Inputaware is particularly significant, with an average improvement of \todo{36.18\%} in ASuR, while in the case of TrojanNN, the enhancement is comparatively limited, with an average improvement of \todo{1.41\%} in ASuR. From the perspective of defense, \tool generally enhances the resistance against various defenses. Significant enhancement in resistance is observed for FP, NAD and NC, with all three exhibiting an average improvement around \todo{27\%} in ASuR.

Of all the four tasks, \tool exhibits the best enhancement on TrojanNN but the most limited enhancement on SSBA, with the average increase in ASuR by \todo{38.87\%} and \todo{14.30\%} on average.
From the perspective of defenses, \tool exhibits the best enhancement against NC but the most limited enhancement against BNP, with the average increase in ASuR by \todo{37.58\%} and \todo{5.31\%} on average.~We observe that most original attacks already have significant resistance against BNP, leaving limited space for improvement.

\section{Results on Face Recognition Task}\label{Appendix:other_tasks}

\begin{table}[!t]
    \caption{Attack survivability comparison between BadNets and \tool-BadNets on InceptionResnetV1 using CFP.} 
    \label{tbl:facenet-survivability}
    \vspace{-10pt}
    \footnotesize
    \centering
    \begin{tabular}{c*{6}{p{0.85cm}}}
        \toprule
        \multirow{2}{*}{Metric} & \multicolumn{2}{c}{No Defense} & \multicolumn{2}{c}{FP} & \multicolumn{2}{c}{FT} \\ 
         \cmidrule(lr){2-3}\cmidrule(lr){4-5} \cmidrule(lr){6-7}
         & O. & V. & O. & V. & O. & V.\\ 
        \midrule 
        BA   & 92.00 & 91.53 & 90.00 & 88.73 & 95.00 & 92.60 \\
        ASR  & 83.00 & 87.71 &11.00 & 75.82 & 3.00  & 41.22 \\
        ASuR & - & - & 12.67 & 73.98 & 6.46  & 42.16 \\
        \bottomrule 
    \end{tabular}
\end{table}

In addition to image classification tasks, we also apply \tool to the face recognition task, where attackers confuse the model~to~recognize faces as a specific person and cause great harm. We select the InceptionResnetV1 model~\cite{szegedy2017inception} and the CFP dataset~\cite{sengupta2016frontal}. CFP~has~500 classes, and 5,000 and 2,000 training and testing samples of a size of 512$\times$512. InceptionResnetV1 has 23.74M parameters, and achieves a benign accuracy of 100.00\% on CFP. Our preliminary evaluation uses \tool to enhance the attack survivability of BadNets against two defenses, i.e, FP and FT. Specifically, in the binary-task training step, we choose \textit{repeat\_3.4.conv2d} as the target layer to generate TCDP, and other settings follow the default one in Appendix~\ref{Appendix:Attack Setting}.

Table~\ref{tbl:survivability} reports the attack survivability comparison results. The original BadNets attack has little survivability against both defenses; i.e., ASR decreases from \todo{83.00\%} to \todo{11.00\%} and \todo{3.00\%} respectively after FP and FT are applied, leading to a low ASuR of \todo{12.67\%} and \todo{6.46\%} respectively. As CFP contains 500 person, with only 14 images for each person, it is more prone to catastrophic forgetting, making FP and FT achieve a good defense capability against BadNets.~However, the \tool-enhanced BadNets still maintains a high ASR of \todo{75.82\%} and \todo{41.22\%} respectively, and exhibits an improved ASuR of \todo{73.98\%} and \todo{42.16\%} respectively against the two defenses.~This~preliminary evaluation result has demonstrated the potential generalizability of \tool to other tasks.

\section{Ethics and Data Privacy}

The training data used in our experiments are all from publicly~available sources, serving solely for research purposes. All the experiments were conducted in a controlled~environment, ensuring that no backdoored models were disseminated to model markets.

\end{document}